\PassOptionsToPackage{table}{xcolor}
\PassOptionsToPackage{capitalise}{cleveref}

\documentclass[mlmain,onecolumn,cleveref]{jmlr}
\jmlrpmlr

\usepackage{booktabs}
\usepackage{multirow}
\usepackage{array}
\usepackage{float}
\usepackage{mathtools}
\usepackage{pifont}
\usepackage{changepage}

\definecolor{bandA}{HTML}{A7CDC8}
\definecolor{bandB}{HTML}{D2E6E2}
\definecolor{bandC}{HTML}{86BDB7}
\definecolor{winner}{HTML}{F6DAB8}

\DeclareMathOperator{\arcosh}{arcosh}

\DeclareMathOperator{\expm}{expm}
\DeclareMathOperator{\aper}{aper}
\DeclareMathOperator{\ext}{ext}
\DeclareMathOperator{\huber}{Huber}
\newcommand{\rd}{r_{d}}
\newcommand{\rext}{r_{\ext}}
\newcommand{\raper}{r_{\aper}}
\newcommand{\Rn}{\mathbb{R}^{n}}

\newcommand{\best}[1]{\cellcolor{winner}#1}
\newcommand{\bestend}[1]{\multicolumn{1}{>{\columncolor{winner}[\tabcolsep][0pt]}c@{}}{#1}}
\newcommand{\yes}{\ding{51}}
\newcommand{\no}{\ding{55}}

\jmlrvolume{}
\firstpageno{1}
\editors{List of editors' names}

\jmlryear{2026}
\jmlrworkshop{Symmetry and Geometry in Neural Representations}
\title[How Hyperbolic Are Hyperbolic VLMs?]%
      {How Hyperbolic Are Hyperbolic Vision-Language Models?}

\author{%
  \Name{Freek Byrman\nametag{$^{1}$}} \Email{freek.byrman@student.uva.nl}\\
  \Name{Partha Das\nametag{$^{1,2}$}} \Email{p.das@uva.nl}\\
  \Name{Stijn Verdenius\nametag{$^{2}$}} \Email{stijn@wairforretail.com}\\
  \Name{Pascal Mettes\nametag{$^{1}$}} \Email{p.s.m.mettes@uva.nl}\\
  \addr $^{1}$University of Amsterdam, $^{2}$WAIR
}
\begin{document}

\maketitle

\begin{abstract}
Hyperbolic vision-language models (VLMs) have emerged as a rapidly growing area of research, reporting strong zero-shot performance alongside an interpretable embedding space in which radius orders representations by generality. But how hyperbolic are hyperbolic VLMs in practice? We investigate this at the exponential map, the single operation through which curvature enters an otherwise Euclidean network. We show that across MERU, HyCoCLIP, ARGENT, and PHyCLIP, the exponential map is well approximated by the identity function, i.e., pairwise distances and angles in the pre-map Euclidean space barely change under projection, and removing the map at inference leaves evaluation scores unchanged. Motivated by this observation, we train Euclidean VLMs on CC3M from scratch under the same hierarchy losses. Over roughly the final $90$\% of the training steps, the training statistics are nearly indistinguishable from those of their hyperbolic counterparts, and the resulting models match the evaluation scores and recover the same radial distributions. Our findings establish Euclidean space as a sufficient substrate for hierarchical vision-language learning as it is currently practiced, and motivate new paradigms in which hyperbolic geometry can deliver on its theoretical promise. Code is available in our \href{https://github.com/fbyrman/hyperbolic-vlms}{GitHub}.
\end{abstract}

\begin{keywords}
  Riemannian manifolds, vision-language models, contrastive learning
\end{keywords}

%%%%%%%%%%%%%%%%%%%%%%%%%%%%%%%%%%%%%%%%%%%%%%%%%%%%%%%%%%%%%%%%%%%%%%%%%%%%%%%
%% Body
%%%%%%%%%%%%%%%%%%%%%%%%%%%%%%%%%%%%%%%%%%%%%%%%%%%%%%%%%%%%%%%%%%%%%%%%%%%%%%%

\section{Introduction}
\label{sec:introduction}

Vision and language are inherently hierarchical. Concepts form tree-like structures within and across modalities: animal subsumes dog subsumes corgi, an image crop abstracts a more detailed scene, and a caption abstracts over every image it could describe. Because trees branch exponentially with depth, they fit poorly in Euclidean space \citep{matouek_1999}, where volume grows only polynomially with radius. Hyperbolic space, whose volume grows exponentially, is mathematically a natural host for such hierarchies \citep{gromov1987}.

On the other hand, contrastive vision-language learning owes none of its early success to this idea: CLIP \citep{CLIP} and ALIGN \citep{ALIGN} showed that angular contrastive learning over hundreds of millions of image-text pairs produces embeddings with strong zero-shot performance. Hyperbolic vision-language models (VLMs) retain this contrastive framework on a hyperbolic manifold and add an entailment loss \citep{entailment_cones}, that orders representations by generality along the radius. MERU \citep{meru} introduced the design, HyCoCLIP \citep{hycoclip} extended it to compositional and intra-modal hierarchies, and variants have followed \citep{atmg, HySAC, ARGENT, phyclip}. However, these rest on Euclidean backbones and enter the manifold in the final layer through an exponential map.

Strong zero-shot classification and retrieval performance rapidly propelled and expanded this line of work. An interpretable and radially organized embedding space, adds to its benefits, fueling efforts on refined loss functions and better benchmark performances.  Yet being hyperbolic is a matter of degree. Curvature sets a characteristic length scale, and how much of the geometry is realized depends on where representations lie relative to it. Since curvature and embeddings are both learned, the effective degree of hyperbolicity is free to vary during training. How much of that geometry hyperbolic VLMs actually use is therefore central to explaining their success, yet this has gone largely uninvestigated.

We investigate this question at the exponential map, the single point at which curvature is introduced in an otherwise Euclidean network. Across MERU \citep{meru}, HyCoCLIP \citep{hycoclip}, ARGENT \citep{ARGENT}, and PHyCLIP \citep{phyclip} trained on CC3M, we observe that this map acts as an identity function: removing it at inference leaves evaluation scores unchanged. Motivated by this observation, we train Euclidean VLMs with the same hierarchy losses from scratch. After the first tenth of the training steps the curvature reaches its lower bound with the training statistics becoming indistinguishable from those of the respective hyperbolic counterparts. Consequently, the Euclidean models achieve the same evaluation scores and reproduce the same radial distributions.
Our findings suggest that the optimization of hyperbolic VLM leaves much to be desired, since none of the state-of-the-art models make use of the part of the space that makes hyperbolic geometry unique and hierarchical.
Our contributions are:

\begin{enumerate}
    \item \textbf{Distortion ratios}: We introduce distortion ratios as a way to quantify hyperbolicity by comparing elementary geometric operations before and after the exponential map.
    \item \textbf{Diagnosis}: We diagnose flagship hyperbolic VLMs as weakly hyperbolic\footnote{\citet{kim2026} independently and concurrently reach a similar conclusion through different but complementary experiments; we discuss the differences between the two works in \cref{sec:discussion}.}: distortion ratios stay near $1$, and exponential-map readout ablations cost no performance.
    \item \textbf{Euclidean baselines}: We introduce Euclidean VLMs with hierarchy losses that exhibit surprisingly similar training dynamics and performance to hyperbolic ones.
\end{enumerate}

\section{Background}
\label{sec:background}

\subsection{Geometry}
\label{sec:geometry}

\paragraph{Euclidean space.} Euclidean space $\mathbb{E}^{n}$ is the vector space $\mathbb{R}^{n}$ equipped with the inner product $\langle \boldsymbol{x}, \boldsymbol{y}\rangle_{\mathbb{E}} \coloneqq \sum_{i=1}^{n} x_i y_i$, where $\boldsymbol{x} = [x_1, \dots, x_n]$ and $\boldsymbol{y} = [y_1, \dots, y_n]$. This inner product induces the norm $\lVert\boldsymbol{x}\rVert_{\mathbb{E}} \coloneqq \sqrt{\langle \boldsymbol{x}, \boldsymbol{x}\rangle_{\mathbb{E}}}$, and the origin of $\mathbb{E}^{n}$ is the vector $\boldsymbol{0}_{\mathbb{E}} \coloneqq (0, \dots, 0)$. Its curvature is zero everywhere, and geodesics are straight segments.

\paragraph{Hyperbolic space.} We use the Lorentz model throughout, which is the common choice in hyperbolic VLM literature ~\citep{meru, hycoclip}. A vector decomposes into space and time components as $\boldsymbol{x} = [\boldsymbol{x}_{\mathrm{space}}, x_{\mathrm{time}}]$ and carries the Lorentzian inner product
\begin{equation}\label{eq:lorentz_inner}
  \langle \boldsymbol{x}, \boldsymbol{y}\rangle_{\mathbb{L}} \coloneqq \langle \boldsymbol{x}_{\mathrm{space}}, \boldsymbol{y}_{\mathrm{space}}\rangle_{\mathbb{E}} - x_{\mathrm{time}}\, y_{\mathrm{time}}.
\end{equation}
The points of fixed Lorentzian norm $-1/c$, for $c > 0$, form a hyperboloid of two sheets; the model keeps the upper one, selected by $x_{\mathrm{time}} > 0$, with origin $\boldsymbol{0}_{\mathbb{L}} \coloneqq [\boldsymbol{0}_{\mathbb{E}}, \sqrt{1/c}]$,
\begin{equation}\label{eq:hyperboloid}
  \mathbb{L}^{n}_{c} \coloneqq \bigl\{\boldsymbol{x} \in \mathbb{R}^{n+1} : \langle \boldsymbol{x}, \boldsymbol{x}\rangle_{\mathbb{L}} = -1/c,\; x_{\mathrm{time}} > 0\bigr\}.
\end{equation}
That sheet has constant negative curvature $-c$ at every point, and the volume of a ball on it grows exponentially with its radius rather than polynomially, as fast as a tree branches.

\paragraph{Core operations.} Almost every training objective in the hyperbolic VLM literature is built from three geometric quantities. The distance between two points is the geodesic length between them; the exterior angle at $\boldsymbol{x}$ toward $\boldsymbol{y}$ is measured from the outward continuation of the ray through $\boldsymbol{x}$; and the half-aperture at $\boldsymbol{x}$ is the angle subtended by a ball around the origin that the cone cannot enter, with radius $R>0$ in the Euclidean case and radius $K>0$ in the hyperbolic case. We introduce them here without subscripts,
\begin{align}
  d(\boldsymbol{x}, \boldsymbol{y}) &\coloneqq
  \begin{dcases}
    \lVert \boldsymbol{x} - \boldsymbol{y}\rVert_{\mathbb{E}}, & \text{in } \mathbb{E}^{n},\\
    \frac{1}{\sqrt{c}}\arcosh\!\bigl(-c\,\langle \boldsymbol{x}, \boldsymbol{y}\rangle_{\mathbb{L}}\bigr), & \text{in } \mathbb{L}^{n}_{c},
  \end{dcases} \label{eq:distance}\\[0.6ex]
  \ext(\boldsymbol{x}, \boldsymbol{y}) &\coloneqq
  \begin{dcases}
    \arccos\!\left(\frac{\lVert\boldsymbol{y}\rVert_{\mathbb{E}}^{2} - \lVert\boldsymbol{x}\rVert_{\mathbb{E}}^{2} - \lVert\boldsymbol{x} - \boldsymbol{y}\rVert_{\mathbb{E}}^{2}}{2\,\lVert\boldsymbol{x}\rVert_{\mathbb{E}} \cdot \lVert\boldsymbol{x} - \boldsymbol{y}\rVert_{\mathbb{E}}}\right), & \text{in } \mathbb{E}^{n},\\
    \arccos\!\left(\frac{y_{\mathrm{time}} + x_{\mathrm{time}}\, c\, \langle \boldsymbol{x}, \boldsymbol{y}\rangle_{\mathbb{L}}}{\lVert \boldsymbol{x}_{\mathrm{space}}\rVert_{\mathbb{E}} \sqrt{\bigl(c\, \langle \boldsymbol{x}, \boldsymbol{y}\rangle_{\mathbb{L}}\bigr)^{2} - 1}}\right), & \text{in } \mathbb{L}^{n}_{c},
  \end{dcases} \label{eq:exterior_angle}\\[0.6ex]
  \aper(\boldsymbol{x}) &\coloneqq
  \begin{dcases}
    \arcsin\!\left(\frac{R}{\lVert\boldsymbol{x}\rVert_{\mathbb{E}}}\right), & \text{in } \mathbb{E}^{n},\\
    \arcsin\!\left(\frac{2K}{\sqrt{c}\,\lVert \boldsymbol{x}_{\mathrm{space}}\rVert_{\mathbb{E}}}\right), & \text{in } \mathbb{L}^{n}_{c}.
  \end{dcases} \label{eq:aperture}
\end{align}
\paragraph{From encoder outputs to the hyperboloid.} Hyperbolic VLMs operate on top of Euclidean backbones. The encoder output $\boldsymbol{v} \in \Rn$ lifts into the tangent space at $\boldsymbol{0}_{\mathbb{L}}$ as $[\boldsymbol{v}, 0]$, and the exponential map carries it a distance $\lVert\boldsymbol{v}\rVert_{\mathbb{E}}$ to
\begin{equation}\label{eq:expmap_origin}
  \expm_{\boldsymbol{0}_{\mathbb{L}}}(\boldsymbol{v}) \coloneqq \left[\frac{\sinh(\sqrt{c}\,r)}{\sqrt{c}\,r}\,\boldsymbol{v},\; \frac{\cosh(\sqrt{c}\,r)}{\sqrt{c}}\right].
\end{equation}

\subsection{Loss functions}
\label{sec:loss_functions}

Two kinds of loss recur across the field: a contrastive term on the distance, and hierarchy terms on the exterior angle and/or the half-aperture. We state them for ordered set embeddings $\mathcal{X} = \{\boldsymbol{x}_i\}_{i=1}^{|\mathcal{B}|}$ and $\mathcal{Y} = \{\boldsymbol{y}_i\}_{i=1}^{|\mathcal{B}|}$ over a batch $\mathcal{B}$, with $\boldsymbol{x}_i$ and $\boldsymbol{y}_i$ as positive pair.

\paragraph{Contrastive loss.} Under similarity $\kappa = -d$ and temperature $\tau$, the softmax loss~\citep{sohn, infonce} aligns positive pairs and repels negative ones,
\begin{equation}\label{eq:softmax_loss}
  \mathcal{L}_{\text{softmax}}(\mathcal{X}, \mathcal{Y}) \coloneqq -\frac{1}{|\mathcal{B}|} \sum_{i=1}^{|\mathcal{B}|} \log \frac{\exp\bigl(\kappa(\boldsymbol{x}_i, \boldsymbol{y}_i)/\tau\bigr)}{\sum_{j=1}^{|\mathcal{B}|}\exp\bigl(\kappa(\boldsymbol{x}_i, \boldsymbol{y}_j)/\tau\bigr)}.
\end{equation}

\paragraph{Hierarchy losses.} The remaining two primitives define an entailment cone~\citep{entailment_cones}: $\boldsymbol{y}$ lies inside the cone at $\boldsymbol{x}$ when $\ext(\boldsymbol{x}, \boldsymbol{y})$ stays below $\eta\,\aper(\boldsymbol{x})$, for a width hyperparameter $\eta$. The hinge is one-sided: pairs inside the cone receive no gradient,
\begin{equation}\label{eq:entailment_loss}
  \mathcal{L}_{\text{ent}}(\mathcal{X}, \mathcal{Y}, \eta) \coloneqq \frac{1}{|\mathcal{B}|}\sum_{i=1}^{|\mathcal{B}|} \max\bigl(0,\; \ext(\boldsymbol{x}_i, \boldsymbol{y}_i) - \eta\,\aper(\boldsymbol{x}_i)\bigr).
\end{equation}
A recent variant~\citep{ARGENT} drops the aperture and penalizes the exterior angle itself, smoothed by a Huber transition at $\beta$ and weighted by a function $h$,
\begin{equation}\label{eq:adaptive_entailment}
  \mathcal{L}_{\text{ada}}(\mathcal{X}, \mathcal{Y}, h) \coloneqq \frac{1}{|\mathcal{B}|}\sum_{i=1}^{|\mathcal{B}|} \huber_{\beta}\bigl(h(\boldsymbol{x}_i, \boldsymbol{y}_i)\,\ext(\boldsymbol{x}_i, \boldsymbol{y}_i)\bigr).
\end{equation}

\section{Methodology}
\label{sec:methodology}

\subsection{Common setup}
\label{sec:setup}

Below we describe our train and evaluation pipeline. The design choices follow standard practice in the hyperbolic vision-language literature, the one difference being a pre-training scale set by our compute budget. \cref{app:setup} provides all hyperparameters and details.

\paragraph{Training.} Every model is trained on CC3M~\citep{CC3M}, of which we retrieve $3{,}016{,}640$ samples, with the grounded boxes released for it by \citet{gligen}. We train each in the ViT-S/16 and ViT-B/16 configurations for $100{,}000$ steps at batch size $512$, once per configuration, as is conventional given the cost of pre-training.

\paragraph{Models.} We consider four models: MERU~\citep{meru}, HyCoCLIP~\citep{hycoclip}, ARGENT~\citep{ARGENT}, and PHyCLIP~\citep{phyclip}. They use two kinds of supervision: image--text pairs, and grounded boxes, i.e., phrases and the image regions they describe. Like CLIP, MERU aligns texts with images via the contrastive loss of \cref{eq:softmax_loss}, and adds the entailment loss of \cref{eq:entailment_loss} so that each image lies in the cone of its text. HyCoCLIP applies both losses to the boxes too, placing each full image and text in the cone of its box. ARGENT and PHyCLIP build on HyCoCLIP: ARGENT uses the adaptive loss of \cref{eq:adaptive_entailment} plus a norm regularizer $\mathcal{L}_{\text{norm}}$, and PHyCLIP splits the embedding space across $64$ low-dimensional hyperboloids, each with its own curvature. We train MERU, HyCoCLIP and PHyCLIP from their public implementations, and implement ARGENT, which has none, ourselves. \cref{app:objectives} gives the exact objectives and weights.

\paragraph{Evaluation.} We report zero-shot classification on sixteen benchmarks, retrieval on COCO~\citep{COCO} and Flickr30K~\citep{Flickr}, and hierarchical classification on the WordNet tree~\citep{Wordnet} over ImageNet~\citep{imagenet}, scored by the path-based metrics TIE and LCA together with the set-based $J$, $P_{\mathrm{H}}$, and $R_{\mathrm{H}}$. Retrieval and classification both rank candidates by the Lorentzian inner product $\langle\cdot,\cdot\rangle_{\mathbb{L}}$.

\subsection{Distortion ratios}
\label{sec:distortion_ratios}

The rapid growth of the hyperbolic VLM literature makes the degree to which these models are actually hyperbolic worth establishing. But how would you even quantify hyperbolicity? Curvature provides the natural starting point. From a global standpoint, any non-zero curvature is meaningfully different from Euclidean space. In practice, however, a model's embeddings do not occupy the entire space; they are confined to a bounded subset of it. Prior work further suggests that this subset is typically small \citep{modality_gap}. Since hyperbolic VLMs introduce curvature through the exponential map, we quantify its practical effect by the ratio of each of the three geometric primitives of \cref{sec:geometry} before and after the map. Those primitives govern every training signal and read-out score.

\paragraph{The three ratios.} Consider two encoder outputs $\boldsymbol{v}_1, \boldsymbol{v}_2 \in \Rn$ with Euclidean norms $r_i \coloneqq \lVert\boldsymbol{v}_i\rVert_{\mathbb{E}}$ and enclosed angle $\theta$, and let $\boldsymbol{x}_i \coloneqq \expm_{\boldsymbol{0}_{\mathbb{L}}}(\boldsymbol{v}_i)$ be their images on $\mathbb{L}^{n}_{c}$. Taking the ratio of the primitives evaluated at these curved and flat coordinates yields three functions:
\begin{equation*}\label{eq:distortion_ratios}
  \rd(r_1, r_2, \theta; c) \coloneqq \frac{d_{\mathbb{L}}(\boldsymbol{x}_1,\boldsymbol{x}_2)}{d_{\mathbb{E}}(\boldsymbol{v}_1,\boldsymbol{v}_2)},
  \quad
  \rext(r_1, r_2, \theta; c) \coloneqq \frac{\ext_{\mathbb{L}}(\boldsymbol{x}_1,\boldsymbol{x}_2)}{\ext_{\mathbb{E}}(\boldsymbol{v}_1,\boldsymbol{v}_2)},
  \quad
  \raper(r_1; c) \coloneqq \frac{\aper_{\mathbb{L}}(\boldsymbol{x}_1)}{\aper_{\mathbb{E}}(\boldsymbol{v}_1)}.
\end{equation*}
\paragraph{Reading the ratio.} For the distance and exterior angle at the converged curvature, sweeping both radii over their empirical distributions and $\theta$ over $[0,\pi]$ defines a three-dimensional volume. We examine this volume one slice at a time, fixing $r_1$ at the mode of one signal’s radial distribution and sweeping the partner radius as a function of the angle. The Euclidean half-aperture introduces a hyperparameter decision on $R$, which we parameterize as $2K/\sqrt{0.1}$. Closed forms for all three ratios are given in \cref{app:derivations}.

\section{Experiments}
\label{sec:experiments}

\subsection{Realized hyperbolicity}
\label{sec:results_hyperbolicity}

\begin{figure}[H]
  \centering
  \includegraphics[width=\textwidth]{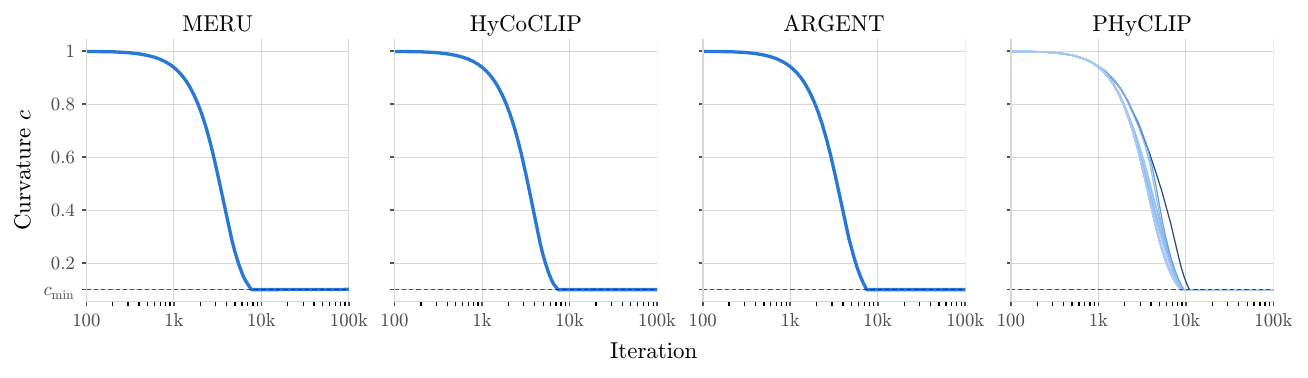}\vspace{-1cm}
  \caption{Curvature $c$ against training iteration for the ViT-S/16 encoders, one panel per objective, with the floor of the admissible range $c_{\min}$ marked. MERU, HyCoCLIP, and ARGENT learn a single curvature; PHyCLIP learns 64 (one per factor).}
  \label{fig:curvature_trajectories}
\end{figure}

\begin{figure}[H]
  \centering
  \includegraphics[width=\textwidth]{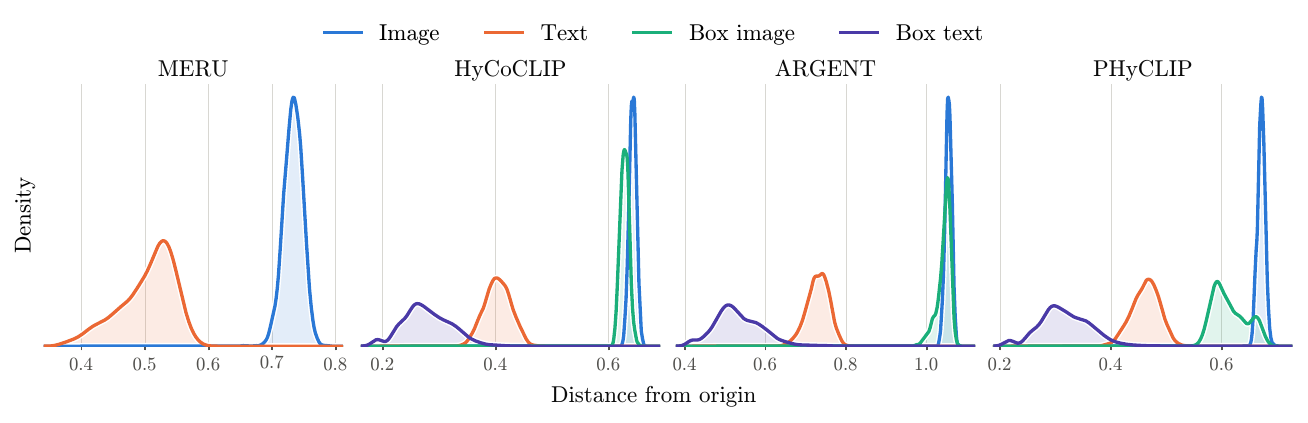}\vspace{-1cm}
  \caption{Radial distribution of the embeddings for the ViT-S/16 models on CC3M. Each panel is a kernel density over distance from the origin, with the horizontal axis scaled per panel. MERU grounds on no boxes and carries only two modalities. For PHyCLIP we visualize the mean of 64 per-factor geodesic distances.}
  \label{fig:radial_distribution}
\end{figure}

We visualize the curvature training trajectories in \cref{fig:curvature_trajectories} for the ViT-S/16 configurations. All models drive their curvature to the lower bound of the admissible range, $c=0.1$, within the first tenth of training. PHyCLIP learns $64$ individual hyperboloid curvatures, all of which likewise reach the clamp. In \cref{fig:radial_distribution}, we visualize the corresponding radial distributions. Together, these figures provide the quantities needed to compute distortion ratio heatmaps.

\vspace{-0.6cm}

\begin{figure}[H]
  \centering
  \includegraphics[width=0.95\textwidth]{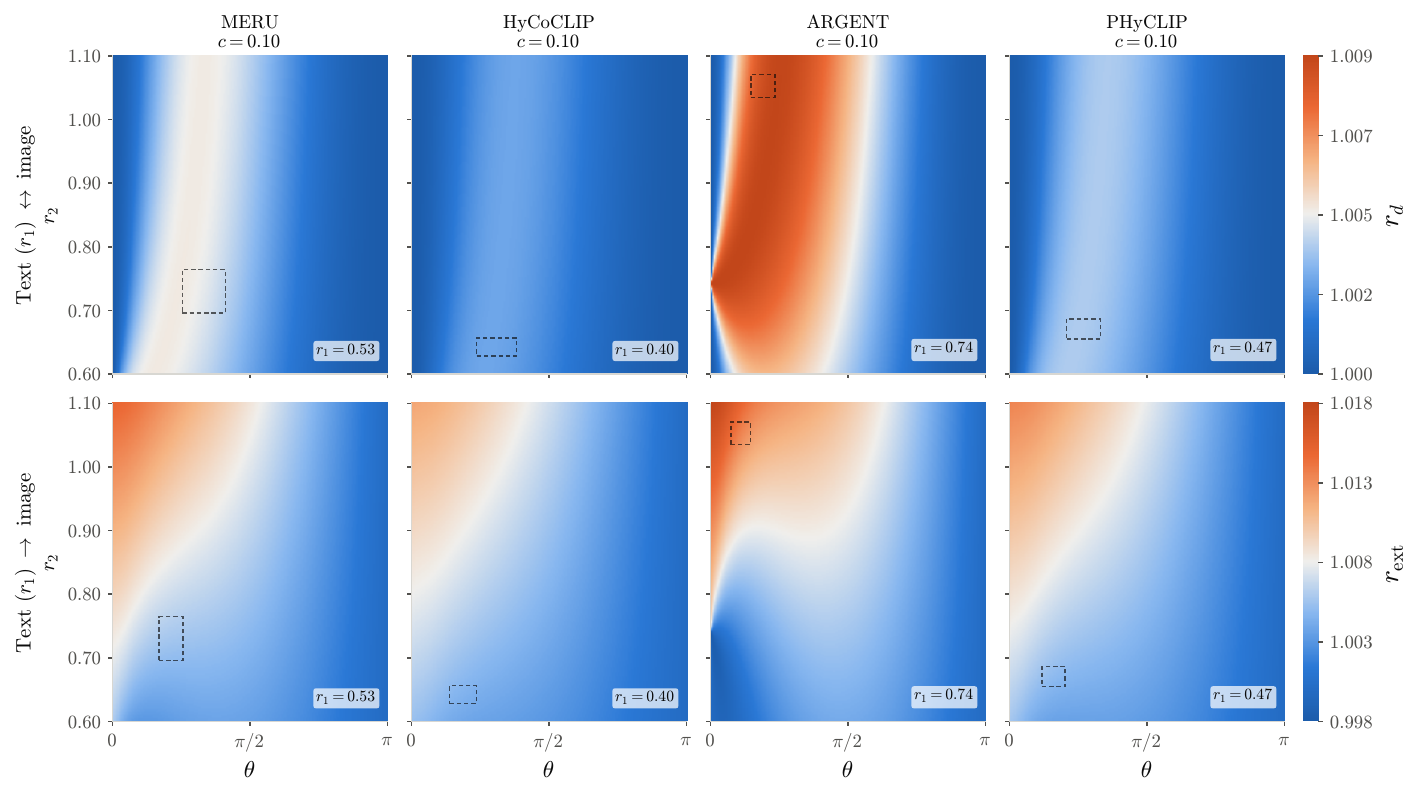}\vspace{-0.7cm}
  \caption{Distance distortion $\rd$ (top) and exterior angle distortion $\rext$ (bottom) for text-image pairs at the converged curvature $c = 0.1$. In each panel $r_1$ is held at the mode of the model's text distribution while the surface sweeps the image radius $r_2$ against the enclosed angle $\theta$; the dashed box marks the region the pairs occupy.}
  \label{fig:distortion_main}
\end{figure}

\Cref{fig:distortion_main} shows distortion ratios for distances and exterior angles in the text--image relation. Across all models, the ratios are close to 1, indicating that Euclidean distances and exterior angles are approximately preserved by the exponential map. Results for the other supervision relations for these geometric quantities are shown in \Cref{fig:distortion_distance_all,fig:distortion_angle_all}.

\begin{figure}[H]
  \centering
  \begin{minipage}[b]{0.65\textwidth}
    \centering
    \includegraphics[width=\textwidth]{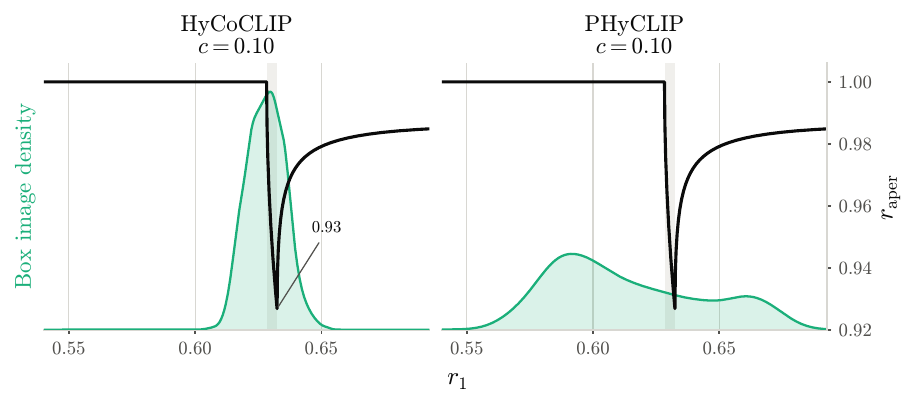}\\[0.1cm]
    {\vspace{-0.35cm}\small (a)}
  \end{minipage}\hfill
  \begin{minipage}[b]{0.33\textwidth}
    \centering
    \includegraphics[width=\textwidth]{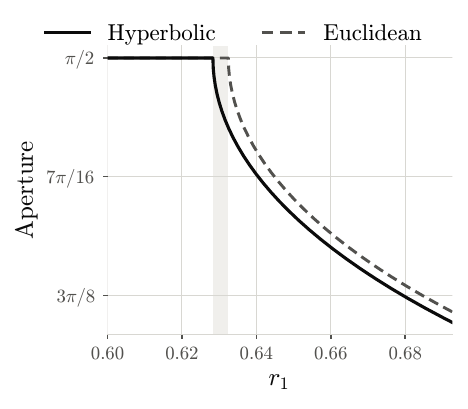}\\[0.1cm]
    {\vspace{-0.35cm}\small (b)}
  \end{minipage}\vspace{-0.35cm}
  \caption{(a) Half-aperture distortion $\raper$ over the box image distribution. Over the shaded interval the flat half-aperture is still clamped at $\pi/2$ while the curved one is not. (b) The two half-apertures themselves over the same radii.}
  \label{fig:aperture_apexes}
\end{figure}

For the remaining primitive, the aperture, the distortion ratio is exactly $1$ for all text and box-text inputs, whose cones are maximally open both before and after the map, a saturation also noted in prior work \citep{ARGENT, kim2026, wang2026}. The ratio at the box-image apexes, shown in \cref{fig:aperture_apexes}(a), is the only one we measure that deviates from $1$. This does not indicate fundamentally different behavior across the two geometries. As \cref{fig:aperture_apexes}(b) shows, the hyperbolic half-aperture un-clamps slightly before the Euclidean one. Because $\arcsin$ is sensitive to inputs near $1$, the ratio briefly dips to $0.93$ before recovering once the Euclidean aperture un-clamps as well. Thus, the two apertures have the same shape, differing only by a small shift in where they leave the clamp.

\subsection{Exponential map ablation at read-out}
\label{sec:euclidean_inference}
If the exponential map is close to the identity function, its removal at read-out and classifying based on Euclidean distances should be undetectable in downstream evaluation scores. \Cref{tab:euclidean_inference} shows that this is indeed the case across all the models we consider: classification and retrieval scores barely move when the exponential map is removed, and even the direction of the movement is inconsistent. Since both readings use the same checkpoint, these differences contain no seed variance. Per-dataset results are given in \Cref{tab:expmap_inference_classification,tab:expmap_inference_retrieval_hierarchical}.\vspace{-0.2cm}

\begin{table}[H]
  \centering
  \caption{Effect of deleting the exponential map at inference. Each checkpoint is read once through $\expm$ and once without it, on tangent-space Euclidean distances. Classification is the mean over sixteen datasets, retrieval over both directions, datasets and cutoffs. $\max_{d}\lvert\Delta\rvert$ is the largest single-benchmark change.}
  \label{tab:euclidean_inference}
  \small
  \setlength{\tabcolsep}{3.8pt}
  \resizebox{\textwidth}{!}{%
  \begin{tabular}{@{}ll cccc cccc@{}}
    \toprule
    & &
    \multicolumn{4}{c}{\cellcolor{bandB}Classification} &
    \multicolumn{4}{c}{\cellcolor{bandA}Retrieval} \\
    \cmidrule{3-10}
    Model & Encoder &
    \cellcolor{bandB!80}With $\expm$ &
    \cellcolor{bandB!80}Without &
    \cellcolor{bandB!80}$\Delta$ &
    \cellcolor{bandB!80}$\max_{d}\lvert\Delta\rvert$ &
    \cellcolor{bandA!80}With $\expm$ &
    \cellcolor{bandA!80}Without &
    \cellcolor{bandA!80}$\Delta$ &
    \multicolumn{1}{>{\columncolor{bandA!80}[\tabcolsep][0pt]}c@{}}{$\max_{d}\lvert\Delta\rvert$} \\
    \midrule
    \multirow{2}{*}{MERU}
    & ViT-S/16
      & 19.66  % classification with expm
      & 19.67  % classification without expm
      & +0.01  % classification Delta
      & 0.20  % classification max |Delta|
      & 44.81  % retrieval with expm
      & 44.83  % retrieval without expm
      & +0.02  % retrieval Delta
      & 0.16  % retrieval max |Delta|
      \\
    & ViT-B/16
      & 22.22  % classification with expm
      & 22.23  % classification without expm
      & +0.01  % classification Delta
      & 0.19  % classification max |Delta|
      & 49.20  % retrieval with expm
      & 49.24  % retrieval without expm
      & +0.03  % retrieval Delta
      & 0.26  % retrieval max |Delta|
      \\
    \midrule
    \multirow{2}{*}{HyCoCLIP}
    & ViT-S/16
      & 23.80  % classification with expm
      & 23.78  % classification without expm
      & -0.02  % classification Delta
      & 0.11  % classification max |Delta|
      & 50.09  % retrieval with expm
      & 50.11  % retrieval without expm
      & +0.02  % retrieval Delta
      & 0.10  % retrieval max |Delta|
      \\
    & ViT-B/16
      & 25.02  % classification with expm
      & 25.01  % classification without expm
      & -0.02  % classification Delta
      & 0.06  % classification max |Delta|
      & 52.47  % retrieval with expm
      & 52.54  % retrieval without expm
      & +0.07  % retrieval Delta
      & 0.30  % retrieval max |Delta|
      \\
    \midrule
    \multirow{2}{*}{ARGENT}
    & ViT-S/16
      & 22.99  % classification with expm
      & 22.95  % classification without expm
      & -0.05  % classification Delta
      & 0.42  % classification max |Delta|
      & 48.34  % retrieval with expm
      & 48.24  % retrieval without expm
      & -0.10  % retrieval Delta
      & 0.70  % retrieval max |Delta|
      \\
    & ViT-B/16
      & 24.35  % classification with expm
      & 24.32  % classification without expm
      & -0.03  % classification Delta
      & 0.39  % classification max |Delta|
      & 52.57  % retrieval with expm
      & 52.56  % retrieval without expm
      & -0.01  % retrieval Delta
      & 0.10  % retrieval max |Delta|
      \\
    \midrule
    \multirow{2}{*}{PHyCLIP}
    & ViT-S/16
      & 23.03  % classification with expm
      & 23.01  % classification without expm
      & -0.03  % classification Delta
      & 0.37  % classification max |Delta|
      & 49.34  % retrieval with expm
      & 49.24  % retrieval without expm
      & -0.11  % retrieval Delta
      & 0.70  % retrieval max |Delta|
      \\
    & ViT-B/16
      & 25.06  % classification with expm
      & 24.97  % classification without expm
      & -0.09  % classification Delta
      & 0.54  % classification max |Delta|
      & 52.35  % retrieval with expm
      & 52.58  % retrieval without expm
      & +0.23  % retrieval Delta
      & 0.46  % retrieval max |Delta|
      \\
    \bottomrule
  \end{tabular}%
  }
\end{table}\vspace{-0.5cm}

\subsection{Exponential map ablation during training}
\label{sec:euclidean_training}

The previous section established that the exponential map can be removed at inference without degrading performance. This does not, however, imply that curvature played no role in shaping the representations during training. To test this, we train a \emph{Euclidean control} for each hyperbolic model using the same objective from scratch, with every operation in \cref{sec:geometry} interpreted in $\mathbb{E}^{n}$ and the exponential map replaced by the identity. Loss components, weights, pairings, data, and schedule are held fixed, so each model and its control differ only in the underlying manifold. We set the ball radius to $R = 2K/\sqrt{0.1}$, matching the Euclidean aperture to the hyperbolic aperture at convergence. For $\mathcal{L}_{\text{norm}}$, we use the full norm in place of the spatial norm, since there is no time component to exclude.

\begin{figure}[H]
  \centering
  \includegraphics[width=\textwidth]{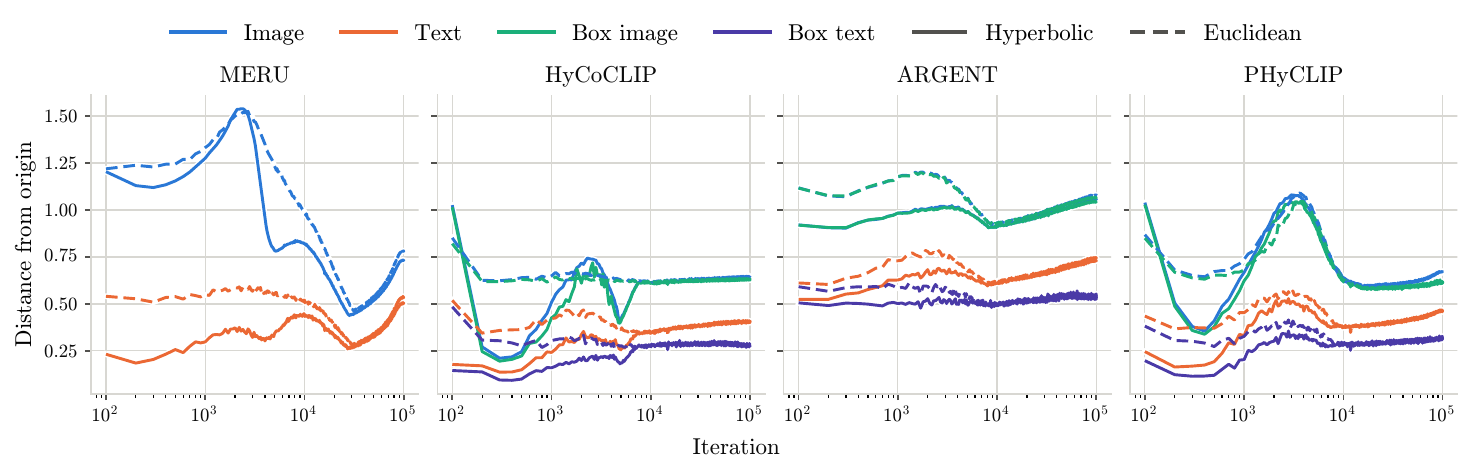}\vspace{-1cm}
  \caption{Mean distance from the origin against training iteration for the ViT-S/16 encoders, one panel per objective and one curve per modality and box type, hyperbolic (solid) against Euclidean control (dashed). MERU grounds on no boxes.}
  \label{fig:origin_distance_trajectories}
\end{figure}

\Cref{fig:origin_distance_trajectories} shows that, after the first 10k steps, the Euclidean and hyperbolic VLMs occupy the same mean radial bands. Likewise, \cref{fig:training_dynamics} shows that their contrastive losses, structuring losses, and temperature evolve to nearly identical magnitudes. The discrepancy during the first 10k iterations arises because the curvature requires gradient updates to reach its lower clamp. Once it does, the training dynamics become effectively indistinguishable. This raises a natural question: does this initial deviation have any lasting effect on final performance?\vspace{-0.1cm}

\vspace{-0.45cm}
We find no evidence that it does: the scores mostly agree with each other. \Cref{tab:euclidean_training_summary} summarizes the evaluation set: on classification, the Euclidean controls win on $6$ of $8$ models in terms of the mean; in retrieval, this extends to $7$ of $8$; and on hierarchical classification, $30$ of the Euclidean metrics do better, $9$ of the hyperbolic ones do better, and $1$ is a tie. We do not read this as the Euclidean models being better: the scores are all clustered close together, and that closeness is itself the point. The exponential map becomes an identity function within the first tenth of training, so for the vast majority of training the two models are optimizing in effectively the same space. Per dataset results are shown in \Cref{tab:euclidean_training_classification,tab:euclidean_training_retrieval_hierarchical}.\vspace{-0.25cm}

\begin{table}[H]
  \centering
  \caption{Each objective in its hyperbolic formulation ($\mathbb{L}_c^n$) against its Euclidean control ($\mathbb{E}^n$). Classification is the mean over the sixteen benchmarks, retrieval over directions, datasets and cutoffs, and the hierarchical metrics are read on WordNet over ImageNet. The \colorbox{winner}{shaded} cell marks the better geometry; ties are unshaded.}\vspace{-0.15cm}
  \label{tab:euclidean_training_summary}
  \small
  \setlength{\tabcolsep}{3pt}
  \resizebox{\textwidth}{!}{%
  \begin{tabular}{@{}ll cc cc cc cc cc cc cc@{}}
    \toprule
    & &
    \multicolumn{2}{c}{\cellcolor{bandB}Classification} &
    \multicolumn{2}{c}{\cellcolor{bandB}Retrieval} &
    \multicolumn{2}{c}{\cellcolor{bandA}TIE ($\downarrow$)} &
    \multicolumn{2}{c}{\cellcolor{bandA}LCA ($\downarrow$)} &
    \multicolumn{2}{c}{\cellcolor{bandA}$J$ ($\uparrow$)} &
    \multicolumn{2}{c}{\cellcolor{bandA}$P_{\mathrm{H}}$ ($\uparrow$)} &
    \multicolumn{2}{>{\columncolor{bandA}[\tabcolsep][0pt]}c@{}}{$R_{\mathrm{H}}$ ($\uparrow$)} \\
    \cmidrule{3-4}\cmidrule{5-6}\cmidrule{7-8}\cmidrule{9-10}\cmidrule{11-12}\cmidrule{13-14}\cmidrule{15-16}
    Model & Encoder &
    \cellcolor{bandB!80}$\mathbb{E}^n$ & \cellcolor{bandB!80}$\mathbb{L}_c^n$ &
    \cellcolor{bandB!80}$\mathbb{E}^n$ & \cellcolor{bandB!80}$\mathbb{L}_c^n$ &
    \cellcolor{bandA!80}$\mathbb{E}^n$ & \cellcolor{bandA!80}$\mathbb{L}_c^n$ &
    \cellcolor{bandA!80}$\mathbb{E}^n$ & \cellcolor{bandA!80}$\mathbb{L}_c^n$ &
    \cellcolor{bandA!80}$\mathbb{E}^n$ & \cellcolor{bandA!80}$\mathbb{L}_c^n$ &
    \cellcolor{bandA!80}$\mathbb{E}^n$ & \cellcolor{bandA!80}$\mathbb{L}_c^n$ &
    \cellcolor{bandA!80}$\mathbb{E}^n$ &
    \multicolumn{1}{>{\columncolor{bandA!80}[\tabcolsep][0pt]}c@{}}{$\mathbb{L}_c^n$} \\
    \midrule
    \multirow{2}{*}{MERU}
    & ViT-S/16 & \best{19.8} & 19.7 & \best{45.7} & 44.8 & \best{6.440} & 6.582 & \best{3.384} & 3.442 & \best{0.605} & 0.596 & \best{0.719} & 0.713 & \best{0.719} & 0.711 \\
    & ViT-B/16 & \best{22.5} & 22.2 & \best{50.0} & 49.2 & \best{6.091} & 6.220 & \best{3.215} & 3.304 & \best{0.626} & 0.621 & \best{0.737} & 0.730 & \best{0.735} & 0.732 \\
    \midrule
    \multirow{2}{*}{HyCoCLIP}
    & ViT-S/16 & 23.5 & \best{23.8} & 47.0 & \best{50.1} & 6.028 & \best{5.851} & 3.180 & \best{3.095} & 0.632 & \best{0.641} & 0.740 & \best{0.748} & 0.738 & \bestend{0.747} \\
    & ViT-B/16 & \best{25.9} & 25.0 & \best{53.3} & 52.5 & \best{5.514} & 5.719 & \best{2.974} & 3.103 & \best{0.664} & 0.652 & \best{0.764} & 0.753 & \best{0.764} & 0.757 \\
    \midrule
    \multirow{2}{*}{ARGENT}
    & ViT-S/16 & 22.7 & \best{23.0} & \best{49.6} & 48.3 & 5.963 & \best{5.931} & \best{3.130} & 3.132 & 0.634 & \best{0.637} & 0.743 & \best{0.746} & 0.740 & \bestend{0.742} \\
    & ViT-B/16 & \best{25.6} & 24.4 & \best{53.0} & 52.6 & \best{5.579} & 5.661 & \best{3.026} & 3.079 & \best{0.661} & 0.656 & \best{0.761} & 0.757 & \best{0.762} & 0.761 \\
    \midrule
    \multirow{2}{*}{PHyCLIP}
    & ViT-S/16 & \best{23.5} & 23.0 & \best{49.8} & 49.3 & \best{5.920} & 5.998 & \best{3.104} & 3.203 & \best{0.636} & 0.633 & \best{0.746} & 0.740 & 0.742 & 0.742 \\
    & ViT-B/16 & \best{25.5} & 25.1 & \best{53.5} & 52.3 & \best{5.588} & 5.745 & \best{3.001} & 3.122 & \best{0.660} & 0.652 & \best{0.760} & 0.752 & \best{0.760} & 0.755 \\
    \bottomrule
  \end{tabular}%
  }
\end{table}

\section{Discussion}\label{sec:discussion}
\paragraph{Relationship to concurrent work.} Independently and concurrently, \citet{kim2026} report that hyperbolic VLMs settle at a near-Euclidean operating point, a conclusion we reach as well using a different set of experiments. Both works share the idea that curvature in isolation cannot quantify hyperbolicity. \citet{kim2026} take a more diagnostic route: they measure curvature gradients in the loss components, audit the operating point using $\sqrt{c}\,\lVert\boldsymbol{x}_{\mathrm{space}}\rVert_{\mathbb{E}}$ as a measure of hyperbolicity, and remove the curvature lower bound to see how this changes convergence behavior. Ours is more interventional: we ask what the exponential map does to geometric operations, then remove it outright, both at read-out on trained checkpoints and by training a Euclidean control from scratch.

\paragraph{Scale.} A natural objection is that CC3M is small compared to the corpora hyperbolic VLMs are trained on. Our arguments, however, rest on relative training dynamics, and we hypothesize that a larger corpus or longer schedule would only extend the near-Euclidean regime rather than escape it. We also repeat the exponential-map removal on the released MERU and HyCoCLIP checkpoints (ARGENT and PHyCLIP have not released weights), trained at much larger scale, and find the map equally inert (\cref{tab:released_classification,tab:released_retrieval_hierarchical}).

\section{Conclusion}
\label{sec:conclusion}

We investigate whether the curvature in current hyperbolic VLMs does the geometric work it is credited for, and find that it does not. The exponential map reduces to an identity function early in training, leaving the space effectively Euclidean: deleting it at read-out leaves scores unchanged, and Euclidean controls trained from scratch match the hyperbolic models' performance and radii closely.  Hyperbolic VLMs have been shown empirically to be effective. We believe that with the right optimization, hyperbolic VLMs can reach much greater performance, which can be measured with our diagnostic tools.

%%%%%%%%%%%%%%%%%%%%%%%%%%%%%%%%%%%%%%%%%%%%%%%%%%%%%%%%%%%%%%%%%%%%%%%%%%%%%%%
%% Bibliography
%%%%%%%%%%%%%%%%%%%%%%%%%%%%%%%%%%%%%%%%%%%%%%%%%%%%%%%%%%%%%%%%%%%%%%%%%%%%%%%

\clearpage
\acks{Computations were performed using the University of Amsterdam - Science Faculty (UvA/FNWI) High Performance Computing Facility, a centrally managed computational resource available to UvA/FNWI researchers including faculty, staff, students, and collaborators.}

\bibliography{references}

%%%%%%%%%%%%%%%%%%%%%%%%%%%%%%%%%%%%%%%%%%%%%%%%%%%%%%%%%%%%%%%%%%%%%%%%%%%%%%%
%% Appendices
%%%%%%%%%%%%%%%%%%%%%%%%%%%%%%%%%%%%%%%%%%%%%%%%%%%%%%%%%%%%%%%%%%%%%%%%%%%%%%%

\appendix
\crefalias{section}{appendix}
\crefalias{subsection}{appendix}
\crefalias{subsubsection}{appendix}

\newpage
\section{Experimental details}
\label{app:setup}

\paragraph{Objectives.}\label{app:objectives} A batch is $\mathcal{B} = \{(\boldsymbol{t}_i, \boldsymbol{i}_i, \boldsymbol{t}_i^{\mathrm{box}}, \boldsymbol{i}_i^{\mathrm{box}})\}_{i=1}^{|\mathcal{B}|}$, in which $\boldsymbol{t}_i$ and $\boldsymbol{i}_i$ are the text and image embeddings of the $i$-th pair and $\boldsymbol{t}_i^{\mathrm{box}}$, $\boldsymbol{i}_i^{\mathrm{box}}$ their grounded-box variants, collected into sets $\mathcal{T}$, $\mathcal{I}$, $\mathcal{T}^{\mathrm{box}}$, and $\mathcal{I}^{\mathrm{box}}$. We train MERU without grounded-box supervision. The contrastive terms are
\begingroup\small
\begin{adjustwidth}{-0.5cm}{-0.5cm}
\begin{align*}
  \mathcal{L}_{\text{softmax}}^{\text{inter}} &\coloneqq \tfrac{1}{2}\bigl(\mathcal{L}_{\text{softmax}}(\mathcal{T}, \mathcal{I}) + \mathcal{L}_{\text{softmax}}(\mathcal{I}, \mathcal{T})\bigr),\\
  \mathcal{L}_{\text{softmax}}^{\text{all}} &\coloneqq \tfrac{1}{4}\bigl(\mathcal{L}_{\text{softmax}}(\mathcal{T}, \mathcal{I}) + \mathcal{L}_{\text{softmax}}(\mathcal{I}, \mathcal{T}) + \mathcal{L}_{\text{softmax}}(\mathcal{T}^{\mathrm{box}}, \mathcal{I}) + \mathcal{L}_{\text{softmax}}(\mathcal{I}^{\mathrm{box}}, \mathcal{T})\bigr),
\end{align*}
\end{adjustwidth}
\endgroup
and every model carries one of the following hierarchy terms alongside them,
\begingroup\small
\begin{adjustwidth}{-0.5cm}{-0.5cm}
\begin{align*}
  \mathcal{L}_{\text{ent}}^{\text{inter}} &\coloneqq \mathcal{L}_{\text{ent}}(\mathcal{T}, \mathcal{I}, \eta_{\text{inter}}),\\
  \mathcal{L}_{\text{ent}}^{\text{all}} &\coloneqq \mathcal{L}_{\text{ent}}(\mathcal{T}^{\mathrm{box}}, \mathcal{I}^{\mathrm{box}}, \eta_{\text{inter}}) + \mathcal{L}_{\text{ent}}(\mathcal{T}, \mathcal{I}, \eta_{\text{inter}}) + \mathcal{L}_{\text{ent}}(\mathcal{I}^{\mathrm{box}}, \mathcal{I}, \eta_{\text{intra}}) + \mathcal{L}_{\text{ent}}(\mathcal{T}^{\mathrm{box}}, \mathcal{T}, \eta_{\text{intra}}),\\
  \mathcal{L}_{\text{ada}}^{\text{all}} &\coloneqq \mathcal{L}_{\text{ada}}(\mathcal{T}^{\mathrm{box}}, \mathcal{I}^{\mathrm{box}}, h_{\text{inter}}) + \mathcal{L}_{\text{ada}}(\mathcal{T}, \mathcal{I}, h_{\text{inter}}) + \mathcal{L}_{\text{ada}}(\mathcal{I}^{\mathrm{box}}, \mathcal{I}, h_{\text{intra}}) + \mathcal{L}_{\text{ada}}(\mathcal{T}^{\mathrm{box}}, \mathcal{T}, h_{\text{intra}}).
\end{align*}
\end{adjustwidth}
\endgroup
The adaptive objective adds a term on the spatial norms, holding them in a band,
\begingroup\small
\begin{equation*}
  \mathcal{L}_{\text{norm}} \coloneqq \frac{1}{4|\mathcal{B}|}\sum_{i=1}^{|\mathcal{B}|}\;\sum_{\boldsymbol{x} \in \{\boldsymbol{t}_i,\, \boldsymbol{i}_i,\, \boldsymbol{t}_i^{\mathrm{box}},\, \boldsymbol{i}_i^{\mathrm{box}}\}} \Bigl(\lVert\boldsymbol{x}_{\mathrm{space}}\rVert_{\mathbb{E}}^{2} - \log\bigl(\lVert\boldsymbol{x}_{\mathrm{space}}\rVert_{\mathbb{E}}\bigr)\Bigr).
\end{equation*}
\endgroup
Each model minimizes a weighted sum of the terms it takes from these blocks,
\begingroup\small
\begin{align*}
  \mathcal{L}_{\text{MERU}} &\coloneqq \mathcal{L}_{\text{softmax}}^{\text{inter}} + 0.2\,\mathcal{L}_{\text{ent}}^{\text{inter}},\\
  \mathcal{L}_{\text{HyCoCLIP}} &\coloneqq \mathcal{L}_{\text{softmax}}^{\text{all}} + 0.1\,\mathcal{L}_{\text{ent}}^{\text{all}},\\
  \mathcal{L}_{\text{ARGENT}} &\coloneqq \mathcal{L}_{\text{softmax}}^{\text{all}} + 0.1\,\mathcal{L}_{\text{ada}}^{\text{all}} + 0.1\,\mathcal{L}_{\text{norm}}.
\end{align*}
\endgroup

PHyCLIP splits the embedding space across $64$ independent hyperboloids of dimension $8$, each with its own learnable curvature. Its contrastive term takes the form of $\mathcal{L}_{\text{softmax}}^{\text{all}}$, but scores a pair by the mean of its geodesic distances over the $64$ factors (normalized $\ell_1$ product metric), and $\mathcal{L}_{\text{ent}}^{\text{all}}$ is imposed inside every factor separately and averaged.

\paragraph{Backbone.} Images are encoded using a Vision Transformer~\citep{vision_transformer} on patches of size $16$, text by the 12-layer, 512-dimensional Transformer~\citep{attention_is_all_you_need} of CLIP~\citep{CLIP}, which tokenizes with byte-pair encoding~\citep{byte_pair_encoding} and truncates at 77 tokens. The shared embedding space is 512-dimensional, except for PHyCLIP, which operates on 64 8-dimensional hyperboloids. Encoder outputs are scaled by the projection scalars $\alpha_{\mathrm{image}}$ and $\alpha_{\mathrm{text}}$ before the exponential map. Images are randomly cropped to $50$--$100\%$ of their area and resized to $224 \times 224$.

\paragraph{Learnable scalars.} Following \citet{meru}, the temperature and the curvature are learned in logarithmic space, as $\log(1/\tau)$ and $\log(c)$. The curvature is initialized at $c = 1$ and clamped to $[0.1, 10]$, except in ARGENT, which clamps to $[0.1, 1]$. The projection scalars $\alpha_{\mathrm{image}}$ and $\alpha_{\mathrm{text}}$ are learned as $\log(\alpha)$ and initialized at $1/\sqrt{512}$.

\paragraph{Loss constants.} The temperature is initialized at $\tau = 0.07$ and clamped below at $\tau = 0.01$. The entailment threshold is $\eta_{\text{inter}} = 1$ for MERU, and $\eta_{\text{inter}} = 0.7$, $\eta_{\text{intra}} = 1.2$ for HyCoCLIP and PHyCLIP. ARGENT replaces it with the adaptive weights $h_{\text{inter}} = 1$ and $h_{\text{intra}} = 1 - \exp\bigl(-d(\boldsymbol{x}, \boldsymbol{y})\bigr)$, under a Huber transition of $\beta = \pi/4$. The aperture constant is $K = 0.1$ wherever a cone is formed; ARGENT forms none.

\paragraph{Optimization.} We use AdamW~\citep{AdamW} with $(\beta_1, \beta_2) = (0.9, 0.98)$. The learning rate warms up linearly over $4{,}000$ steps to $5 \times 10^{-4}$, then decays by cosine to zero~\citep{cosine_decay}. We train in mixed precision~\citep{mixed_precision}, computing the exponential map and all losses in FP32. Weight decay is disabled for gains, biases, and the learnable scalars $\tau$, $c$, $\alpha_{\mathrm{image}}$, and $\alpha_{\mathrm{text}}$. PHyCLIP is the exception, left as published: its per-factor curvatures and projection scalars are not excluded, and decay along with the encoder weights. Both are parameterized logarithmically, so decaying the parameter toward zero pulls the quantity it encodes toward $1$, which for the curvature is a pull away from the floor.

\paragraph{Euclidean controls.} Each control keeps every learnable scalar of its hyperbolic twin except the curvature. In particular the projection scalars $\alpha_{\mathrm{image}}$ and $\alpha_{\mathrm{text}}$ are retained and applied identically; encoder outputs are not $L_2$-normalized in either arm, so this scalar is the only rescaling between the encoder and the loss. At read-out the controls score by $-d$, the quantity their contrastive term trains on. The hyperbolic arms induce the same ordering: their released implementations score by $\langle\cdot,\cdot\rangle_{\mathbb{L}}$, and \cref{eq:distance} is strictly decreasing in it at a fixed curvature. PHyCLIP is the exception, evaluated by the sum of the inner products over its $64$ factors, which is not monotone in the mean of their geodesic distances; we leave it as published.

\paragraph{Retrieval.} We evaluate on COCO~\citep{COCO} val2017 and on the Flickr30K~\citep{Flickr} test split of \citet{test_split_flickr}, which hold $5000$ and $1000$ images respectively, each paired with five captions. We report recall at five and ten in both directions.

\paragraph{Zero-shot classification.} We evaluate on sixteen benchmarks in three groups. General object and scene recognition is covered by ImageNet~\citep{imagenet}, on its validation split, CIFAR-10 and CIFAR-100~\citep{cifar}, SUN397~\citep{sun397}, Caltech-101~\citep{caltech}, and STL-10~\citep{stl10}; fine-grained recognition by Food-101~\citep{food}, CUB~\citep{cub}, Stanford Cars~\citep{cars}, FGVC-Aircraft~\citep{aircraft}, Oxford-IIIT Pets~\citep{pets}, and Oxford Flowers-102~\citep{flowers}; and specialized domains by DTD~\citep{DTD}, EuroSAT~\citep{eurosat}, RESISC45~\citep{resic45}, and Country211~\citep{CLIP}. The class prompts are the per-dataset templates released with \citet{phyclip}, which carry over unchanged from \citet{meru} through \citet{hycoclip}; we hold them fixed across every model and control.

\paragraph{Hierarchical classification.} The ImageNet classes are leaves of the WordNet~\citep{Wordnet} hierarchy, so an error can be scored by how far the prediction $\hat{c}$ lies from the truth $c$ in that tree. The \emph{tree-induced error} (TIE) is the number of edges between $\hat{c}$ and $c$, with unit edge weights as is standard for WordNet; the \emph{lowest-common-ancestor} (LCA) error is the height of the deepest node ancestral to both. Both are lower-is-better. The set-based metrics compare the ancestor sets $A(\hat{c})$ and $A(c)$ through their overlap,
\begin{align}
  J &\coloneqq \frac{\lvert A(\hat{c}) \cap A(c)\rvert}{\lvert A(\hat{c}) \cup A(c)\rvert}, \label{eq:jaccard}\\
  P_{\mathrm{H}} &\coloneqq \frac{\lvert A(\hat{c}) \cap A(c)\rvert}{\lvert A(\hat{c})\rvert}, \label{eq:hier_precision}\\
  R_{\mathrm{H}} &\coloneqq \frac{\lvert A(\hat{c}) \cap A(c)\rvert}{\lvert A(c)\rvert}, \label{eq:hier_recall}
\end{align}
yielding the Jaccard similarity $J$ and the hierarchical precision $P_{\mathrm{H}}$ and recall $R_{\mathrm{H}}$. For all set-based metrics, it holds that higher is better.

\newpage
\section{Derivations}
\label{app:derivations}

\subsection{Distance ratio}
\label{sec:appendix_derivations_distance_distortion}
\paragraph{Setup.}
Let $\boldsymbol{v}_1, \boldsymbol{v}_2 \in \mathbb{R}^n$ be two Euclidean encoder outputs, with norms
$r_i \coloneqq \lVert \boldsymbol{v}_i \rVert_{\mathbb{E}}$ and enclosed angle $\theta \in [0, \pi]$ defined through
\begin{equation}\label{eq:angle_def}
\langle \boldsymbol{v}_1, \boldsymbol{v}_2 \rangle_{\mathbb{E}} = r_1 r_2 \cos\theta.
\end{equation}

\paragraph{Euclidean distance.}
Expanding the Euclidean distance (\cref{eq:distance}) and substituting
\cref{eq:angle_def} recovers the law of cosines,
\begin{equation}\label{eq:euclidean_baseline}
d_{\mathbb{E}}(\boldsymbol{v}_1, \boldsymbol{v}_2)
= \sqrt{\langle \boldsymbol{v}_1 - \boldsymbol{v}_2,\, \boldsymbol{v}_1 - \boldsymbol{v}_2 \rangle_{\mathbb{E}}}
= \sqrt{r_1^2 + r_2^2 - 2 r_1 r_2 \cos\theta}.
\end{equation}

\paragraph{Lift to the tangent space at the origin.}
Augment each encoder output to an ambient vector $\bar{\boldsymbol{v}}_i \coloneqq [\boldsymbol{v}_i, 0] \in \mathbb{R}^{n+1}$.
With the origin $\boldsymbol{0}_{\mathbb{L}} = [\boldsymbol{0}_{\mathbb{E}}, \sqrt{1/c}]$ and the Lorentzian inner product (\cref{eq:lorentz_inner}),
\begin{equation}
\langle \bar{\boldsymbol{v}}_i, \boldsymbol{0}_{\mathbb{L}} \rangle_{\mathbb{L}}
= \langle \boldsymbol{v}_i, \boldsymbol{0}_{\mathbb{E}} \rangle_{\mathbb{E}} - 0 \cdot \sqrt{1/c} = 0,
\end{equation}
so $\bar{\boldsymbol{v}}_i \in \mathcal{T}_{\boldsymbol{0}_{\mathbb{L}}} \mathbb{L}_c^n$. Because its time
component vanishes, the Lorentzian norm of the lifted vector reduces to the Euclidean norm of the original,
\begin{equation}\label{eq:norm_reduction}
\lVert \bar{\boldsymbol{v}}_i \rVert_{\mathbb{L}}
= \sqrt{\lvert \langle \boldsymbol{v}_i, \boldsymbol{v}_i \rangle_{\mathbb{E}} \rvert} = r_i.
\end{equation}

\paragraph{Projection onto the hyperboloid.}
Applying the exponential map at the origin (\cref{eq:expmap_origin}) to $\bar{\boldsymbol{v}}_i$ and
using \cref{eq:norm_reduction},
\begin{equation}
\boldsymbol{x}_i \coloneqq \mathrm{expm}_{\boldsymbol{0}_{\mathbb{L}}}(\bar{\boldsymbol{v}}_i)
= \cosh\!\bigl(\sqrt{c}\, r_i\bigr)\, \boldsymbol{0}_{\mathbb{L}}
+ \frac{\sinh\!\bigl(\sqrt{c}\, r_i\bigr)}{\sqrt{c}\, r_i}\, \bar{\boldsymbol{v}}_i,
\end{equation}
which separates into spatial and temporal components
\begin{equation}\label{eq:mapped_components}
\boldsymbol{x}_{i,\mathrm{space}} = \frac{\sinh\!\bigl(\sqrt{c}\, r_i\bigr)}{\sqrt{c}\, r_i}\, \boldsymbol{v}_i,
\qquad
x_{i,\mathrm{time}} = \frac{\cosh\!\bigl(\sqrt{c}\, r_i\bigr)}{\sqrt{c}}.
\end{equation}

\paragraph{Lorentzian inner product.}
Combining the components of \cref{eq:mapped_components} through \cref{eq:lorentz_inner},
the spatial term carries the angle via \cref{eq:angle_def} and the factors $r_1 r_2$ cancel:
\begin{align}
\langle \boldsymbol{x}_1, \boldsymbol{x}_2 \rangle_{\mathbb{L}}
&= \langle \boldsymbol{x}_{1,\mathrm{space}}, \boldsymbol{x}_{2,\mathrm{space}} \rangle_{\mathbb{E}}
   - x_{1,\mathrm{time}}\, x_{2,\mathrm{time}} \nonumber\\
&= \frac{1}{c}\Bigl[
   \sinh\!\bigl(\sqrt{c}\, r_1\bigr)\sinh\!\bigl(\sqrt{c}\, r_2\bigr)\cos\theta
   - \cosh\!\bigl(\sqrt{c}\, r_1\bigr)\cosh\!\bigl(\sqrt{c}\, r_2\bigr)
   \Bigr].
\end{align}

\paragraph{Geodesic distance.}
Substituting into the Lorentzian distance (\cref{eq:distance}), the prefactor $-c$
cancels the $1/c$ and flips the sign of the bracket:
\begin{equation}\label{eq:geodesic_derived}
d_{\mathbb{L}}(\boldsymbol{x}_1, \boldsymbol{x}_2)
= \frac{1}{\sqrt{c}}\, \operatorname{arcosh}\!\Bigl(
   \cosh\!\bigl(\sqrt{c}\, r_1\bigr)\cosh\!\bigl(\sqrt{c}\, r_2\bigr)
   - \sinh\!\bigl(\sqrt{c}\, r_1\bigr)\sinh\!\bigl(\sqrt{c}\, r_2\bigr)\cos\theta
   \Bigr).
\end{equation}

\paragraph{Distance ratio.}
Dividing the geodesic distance in \cref{eq:geodesic_derived} by the Euclidean distance in
\cref{eq:euclidean_baseline} recovers the closed form:
\begin{align}
\rd(r_1, r_2, \theta; c)
&= \frac{d_{\mathbb{L}}(\boldsymbol{x}_1, \boldsymbol{x}_2)}{d_{\mathbb{E}}(\boldsymbol{v}_1, \boldsymbol{v}_2)} \nonumber\\
\label{eq:distortion_ratio}
&= \frac{\operatorname{arcosh}\!\Bigl(\cosh(\sqrt{c}\, r_1)\cosh(\sqrt{c}\, r_2)
        - \sinh(\sqrt{c}\, r_1)\sinh(\sqrt{c}\, r_2)\cos\theta\Bigr)}
       {\sqrt{c}\,\sqrt{r_1^2 + r_2^2 - 2 r_1 r_2 \cos\theta}}.
\end{align}
It is convenient to name the argument of the $\operatorname{arcosh}$, which recurs below,
\begin{equation}\label{eq:A_def}
\begin{split}
A(r_1, r_2, \theta; c) &\coloneqq \cosh\!\bigl(\sqrt{c}\, r_1\bigr)\cosh\!\bigl(\sqrt{c}\, r_2\bigr)
 - \sinh\!\bigl(\sqrt{c}\, r_1\bigr)\sinh\!\bigl(\sqrt{c}\, r_2\bigr)\cos\theta\\
 &= -c\,\langle \boldsymbol{x}_1, \boldsymbol{x}_2 \rangle_{\mathbb{L}}.
\end{split}
\end{equation}

\subsection{Exterior angle ratio}
\label{sec:appendix_derivations_ext_ratio}

\paragraph{Euclidean exterior angle.}
Substituting $\langle \boldsymbol{v}_1, \boldsymbol{v}_2 \rangle_{\mathbb{E}} = r_1 r_2 \cos\theta$ and
$\lVert \boldsymbol{v}_1 \rVert_{\mathbb{E}} = r_1$ into \cref{eq:exterior_angle}, a factor $r_1$ cancels from
numerator and denominator and the Euclidean exterior angle depends on the pair only through $(r_1, r_2, \theta)$,
\begin{equation}\label{eq:ext_euclidean_closed}
\ext_{\mathbb{E}}(\boldsymbol{v}_1, \boldsymbol{v}_2)
= \arccos\!\left(\frac{r_2 \cos\theta - r_1}{\sqrt{r_1^2 + r_2^2 - 2 r_1 r_2 \cos\theta}}\right).
\end{equation}

\paragraph{Lorentzian exterior angle.}
The mapped components \cref{eq:mapped_components} give $x_{i,\mathrm{time}} = \cosh(\sqrt{c}\,r_i)/\sqrt{c}$ and
$\lVert \boldsymbol{x}_{1,\mathrm{space}} \rVert_{\mathbb{E}} = \sinh(\sqrt{c}\,r_1)/\sqrt{c}$, while
\cref{eq:A_def} gives $c\,\langle \boldsymbol{x}_1, \boldsymbol{x}_2 \rangle_{\mathbb{L}} = -A$. Substituting these
into \cref{eq:exterior_angle}, the numerator becomes
$\bigl(\cosh(\sqrt{c}\,r_2) - A\cosh(\sqrt{c}\,r_1)\bigr)/\sqrt{c}$ and the denominator
$\sinh(\sqrt{c}\,r_1)\sqrt{A^2 - 1}/\sqrt{c}$, so the factors of $\sqrt{c}$ cancel and
\begin{equation}\label{eq:ext_lorentz_closed}
\ext_{\mathbb{L}}(\boldsymbol{x}_1, \boldsymbol{x}_2)
= \arccos\!\left(\frac{\cosh\!\bigl(\sqrt{c}\, r_2\bigr) - A\,\cosh\!\bigl(\sqrt{c}\, r_1\bigr)}
                      {\sinh\!\bigl(\sqrt{c}\, r_1\bigr)\,\sqrt{A^2 - 1}}\right).
\end{equation}

\paragraph{Exterior angle ratio.}
Dividing \cref{eq:ext_lorentz_closed} by \cref{eq:ext_euclidean_closed},
\begin{equation}\label{eq:ext_ratio}
\rext(r_1, r_2, \theta; c)
= \arccos\!\left(\frac{\cosh(\sqrt{c}\, r_2) - A\,\cosh(\sqrt{c}\, r_1)}
                      {\sinh(\sqrt{c}\, r_1)\,\sqrt{A^2 - 1}}\right)
  \Bigg/
  \arccos\!\left(\frac{r_2 \cos\theta - r_1}{\sqrt{r_1^2 + r_2^2 - 2 r_1 r_2 \cos\theta}}\right).
\end{equation}

\subsection{Aperture ratio}
\label{sec:appendix_derivations_aper_ratio}

The half-aperture reads a single point, so its ratio is a function of one radius. From \cref{eq:mapped_components},
$\lVert \boldsymbol{x}_{\mathrm{space}} \rVert_{\mathbb{E}} = \sinh(\sqrt{c}\,r_1)/\sqrt{c}$, so the Lorentzian
half-aperture of \cref{eq:aperture} loses its explicit curvature factor,
\begin{equation}\label{eq:aper_lorentz_closed}
\aper_{\mathbb{L}}(\boldsymbol{x})
= \arcsin\!\left(\frac{2K}{\sqrt{c}\,\lVert \boldsymbol{x}_{\mathrm{space}} \rVert_{\mathbb{E}}}\right)
= \arcsin\!\left(\frac{2K}{\sinh\!\bigl(\sqrt{c}\, r_1\bigr)}\right).
\end{equation}
With the ball radius fixed by the matching rule $R = 2K/\sqrt{c}$ of \cref{sec:euclidean_training}, the Euclidean
half-aperture is $\arcsin\bigl(2K/(\sqrt{c}\,r_1)\bigr)$, and taking the ratio curved-over-flat,
\begin{equation}\label{eq:aper_ratio}
\raper(r_1; c) = \frac{\arcsin\bigl(2K/\sinh(\sqrt{c}\, r_1)\bigr)}{\arcsin\bigl(2K/(\sqrt{c}\,r_1)\bigr)}.
\end{equation}

\section{Detailed results}
\label{app:detailed_results}

\begin{figure}[H]
  \centering
  \includegraphics[width=0.8\textwidth]{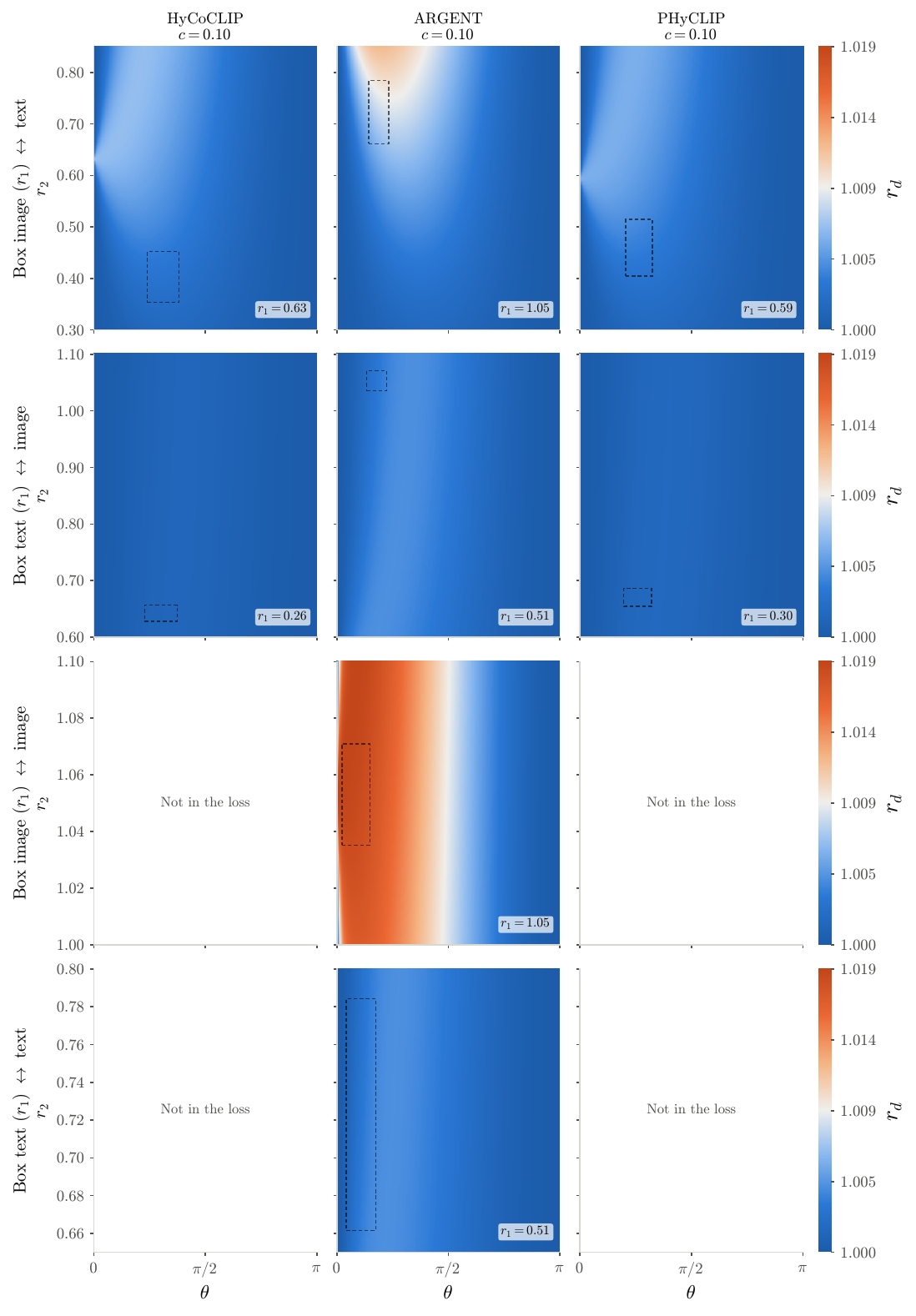}\vspace{-0.5cm}
  \caption{Distance distortion $\rd$ for the remaining pairs which are used in the loss components. In each panel $r_1$ is held at the mode of the signal named first while the surface sweeps the partner's radius $r_2$ against the enclosed angle $\theta$; the dashed box marks the region the pairs occupy. The last two rows are the box-to-source distances ARGENT weights its intra-modal entailment terms by.}
  \label{fig:distortion_distance_all}
\end{figure}

\begin{figure}[H]
  \centering
  \includegraphics[width=0.8\textwidth]{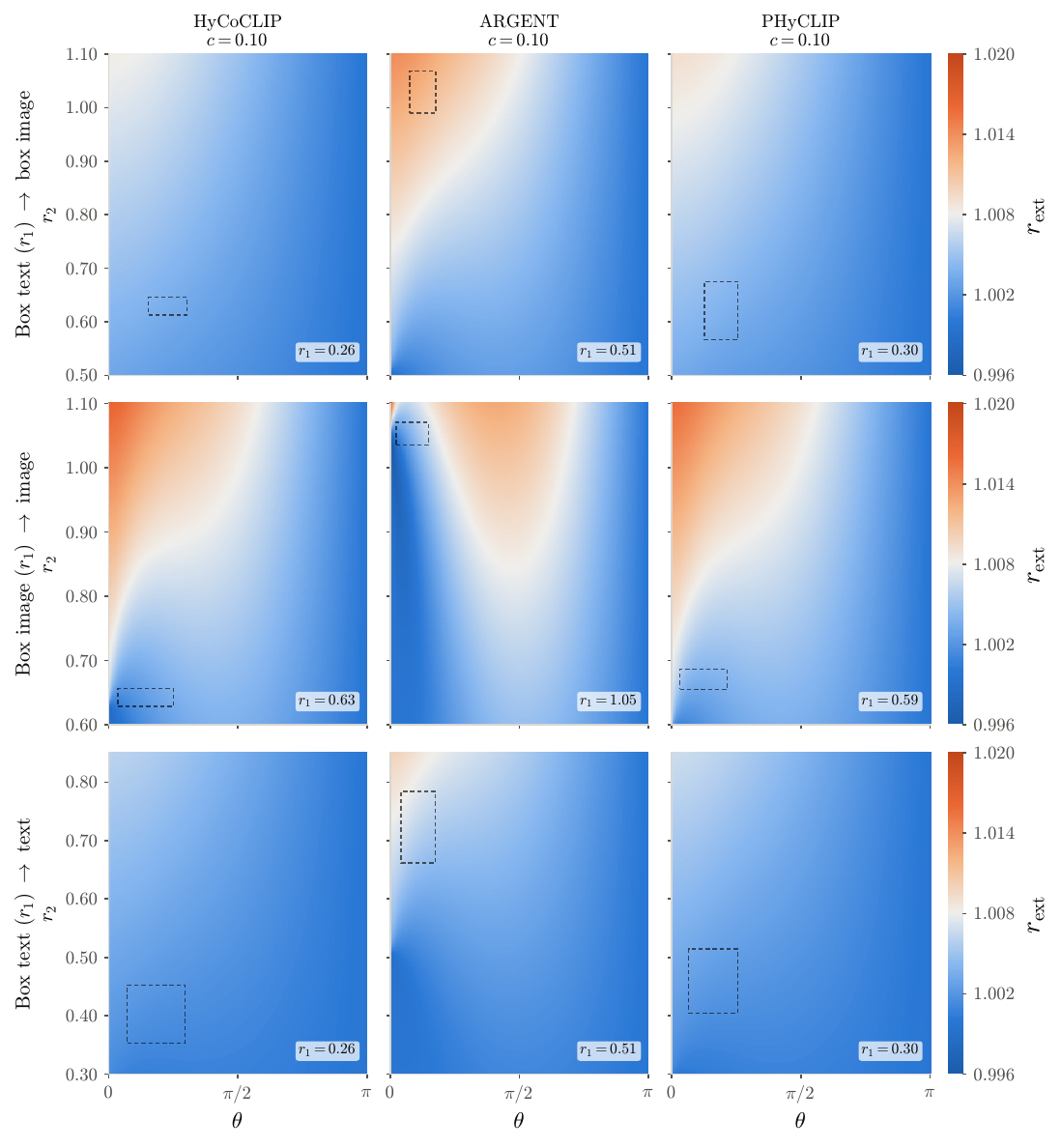}\vspace{-0.5cm}
  \caption{Exterior angle distortion $\rext$ for the remaining pairs which are used in the loss components. In each panel $r_1$ is held at the mode of the signal named first while the surface sweeps the partner's radius $r_2$ against the enclosed angle $\theta$; the dashed box marks the region the pairs occupy.}
  \label{fig:distortion_angle_all}
\end{figure}

\begin{table}[H]
  \centering
  \caption{Zero-shot classification accuracy (\%) across the sixteen benchmarks, for each trained hyperbolic checkpoint read with the exponential map (\yes) and without it (\no). The \colorbox{winner}{shaded} cell marks the better reading in each column, per model and encoder; ties are unshaded.}
  \label{tab:expmap_inference_classification}
  \small
  \setlength{\tabcolsep}{8pt}
  \resizebox{\textwidth}{!}{%
  \begin{tabular}{@{}lllcccccc cccccc cccc}
    \toprule
    & & &
    \multicolumn{6}{c}{\cellcolor{bandB}General} &
    \multicolumn{6}{c}{\cellcolor{bandA}Fine-grained} &
    \multicolumn{4}{c}{\cellcolor{bandC}Specialized} \\
    \cmidrule{4-19}
    Model & Encoder & $\expm$ &
    \cellcolor{bandB!80}\rotatebox{90}{ImageNet} &
    \cellcolor{bandB!80}\rotatebox{90}{CIFAR-10} &
    \cellcolor{bandB!80}\rotatebox{90}{CIFAR-100} &
    \cellcolor{bandB!80}\rotatebox{90}{SUN397} &
    \cellcolor{bandB!80}\rotatebox{90}{Caltech-101} &
    \cellcolor{bandB!80}\rotatebox{90}{STL-10} &
    \cellcolor{bandA!80}\rotatebox{90}{Food-101} &
    \cellcolor{bandA!80}\rotatebox{90}{CUB} &
    \cellcolor{bandA!80}\rotatebox{90}{Cars} &
    \cellcolor{bandA!80}\rotatebox{90}{Aircraft} &
    \cellcolor{bandA!80}\rotatebox{90}{Pets} &
    \cellcolor{bandA!80}\rotatebox{90}{Flowers-102} &
    \cellcolor{bandC!80}\rotatebox{90}{DTD} &
    \cellcolor{bandC!80}\rotatebox{90}{EuroSAT} &
    \cellcolor{bandC!80}\rotatebox{90}{RESISC45} &
    \cellcolor{bandC!80}\rotatebox{90}{Country211} \\
    \midrule
    \multirow{4}{*}{MERU}
    & \multirow{2}{*}{ViT-S/16}
    & \yes
      & 14.3  % ImageNet
      & 53.1  % CIFAR-10
      & 22.9  % CIFAR-100
      & \colorbox{winner}{27.6}  % SUN397
      & \colorbox{winner}{42.1}  % Caltech-101
      & 75.3  % STL-10
      & 9.3  % Food-101
      & 3.0  % CUB
      & 0.9  % Cars
      & 1.2  % Aircraft
      & 9.8  % Pets
      & 7.3  % Flowers-102
      & 8.0  % DTD
      & 18.4  % EuroSAT
      & \colorbox{winner}{20.4}  % RESISC45
      & 0.8  % Country211
      \\
    & & \no
      & 14.3  % ImageNet
      & \colorbox{winner}{53.2}  % CIFAR-10
      & 22.9  % CIFAR-100
      & 27.4  % SUN397
      & 42.0  % Caltech-101
      & \colorbox{winner}{75.5}  % STL-10
      & \colorbox{winner}{9.4}  % Food-101
      & 3.0  % CUB
      & 0.9  % Cars
      & 1.2  % Aircraft
      & 9.8  % Pets
      & \colorbox{winner}{7.4}  % Flowers-102
      & \colorbox{winner}{8.1}  % DTD
      & \colorbox{winner}{18.5}  % EuroSAT
      & 20.3  % RESISC45
      & 0.8  % Country211
      \\
    \cmidrule{2-19}
    & \multirow{2}{*}{ViT-B/16}
    & \yes
      & 16.7  % ImageNet
      & \colorbox{winner}{60.5}  % CIFAR-10
      & 25.9  % CIFAR-100
      & 30.9  % SUN397
      & 47.5  % Caltech-101
      & 84.0  % STL-10
      & 10.7  % Food-101
      & 2.7  % CUB
      & 1.2  % Cars
      & 1.7  % Aircraft
      & 10.9  % Pets
      & 9.0  % Flowers-102
      & 10.8  % DTD
      & 19.1  % EuroSAT
      & 23.1  % RESISC45
      & 0.7  % Country211
      \\
    & & \no
      & 16.7  % ImageNet
      & 60.3  % CIFAR-10
      & 25.9  % CIFAR-100
      & 30.9  % SUN397
      & \colorbox{winner}{47.6}  % Caltech-101
      & \colorbox{winner}{84.1}  % STL-10
      & 10.7  % Food-101
      & 2.7  % CUB
      & 1.2  % Cars
      & 1.7  % Aircraft
      & 10.9  % Pets
      & \colorbox{winner}{9.2}  % Flowers-102
      & \colorbox{winner}{10.9}  % DTD
      & 19.1  % EuroSAT
      & 23.1  % RESISC45
      & 0.7  % Country211
      \\
    \midrule
    \multirow{4}{*}{HyCoCLIP}
    & \multirow{2}{*}{ViT-S/16}
    & \yes
      & 18.4  % ImageNet
      & \colorbox{winner}{72.6}  % CIFAR-10
      & 31.3  % CIFAR-100
      & \colorbox{winner}{33.0}  % SUN397
      & 50.0  % Caltech-101
      & 85.0  % STL-10
      & 10.4  % Food-101
      & \colorbox{winner}{3.6}  % CUB
      & 1.1  % Cars
      & 1.6  % Aircraft
      & 15.1  % Pets
      & 7.8  % Flowers-102
      & 13.5  % DTD
      & \colorbox{winner}{13.1}  % EuroSAT
      & \colorbox{winner}{23.6}  % RESISC45
      & 0.7  % Country211
      \\
    & & \no
      & 18.4  % ImageNet
      & 72.5  % CIFAR-10
      & \colorbox{winner}{31.4}  % CIFAR-100
      & 32.9  % SUN397
      & 50.0  % Caltech-101
      & 85.0  % STL-10
      & 10.4  % Food-101
      & 3.5  % CUB
      & 1.1  % Cars
      & 1.6  % Aircraft
      & 15.1  % Pets
      & 7.8  % Flowers-102
      & 13.5  % DTD
      & 13.0  % EuroSAT
      & 23.5  % RESISC45
      & 0.7  % Country211
      \\
    \cmidrule{2-19}
    & \multirow{2}{*}{ViT-B/16}
    & \yes
      & 19.9  % ImageNet
      & 75.0  % CIFAR-10
      & \colorbox{winner}{34.1}  % CIFAR-100
      & 34.8  % SUN397
      & 50.3  % Caltech-101
      & 88.5  % STL-10
      & 11.7  % Food-101
      & 3.8  % CUB
      & 1.3  % Cars
      & 1.2  % Aircraft
      & 16.2  % Pets
      & 7.6  % Flowers-102
      & 12.0  % DTD
      & 19.1  % EuroSAT
      & 24.0  % RESISC45
      & \colorbox{winner}{0.7}  % Country211
      \\
    & & \no
      & 19.9  % ImageNet
      & 75.0  % CIFAR-10
      & 34.0  % CIFAR-100
      & 34.8  % SUN397
      & 50.3  % Caltech-101
      & \colorbox{winner}{88.6}  % STL-10
      & 11.7  % Food-101
      & 3.8  % CUB
      & 1.3  % Cars
      & 1.2  % Aircraft
      & 16.2  % Pets
      & 7.6  % Flowers-102
      & 12.0  % DTD
      & 19.1  % EuroSAT
      & 24.0  % RESISC45
      & 0.6  % Country211
      \\
    \midrule
    \multirow{4}{*}{ARGENT}
    & \multirow{2}{*}{ViT-S/16}
    & \yes
      & 18.1  % ImageNet
      & \colorbox{winner}{70.7}  % CIFAR-10
      & \colorbox{winner}{28.9}  % CIFAR-100
      & \colorbox{winner}{31.8}  % SUN397
      & \colorbox{winner}{45.2}  % Caltech-101
      & 85.2  % STL-10
      & \colorbox{winner}{9.5}  % Food-101
      & 3.2  % CUB
      & 0.9  % Cars
      & \colorbox{winner}{1.8}  % Aircraft
      & 10.9  % Pets
      & 8.6  % Flowers-102
      & \colorbox{winner}{10.9}  % DTD
      & 17.9  % EuroSAT
      & \colorbox{winner}{23.8}  % RESISC45
      & 0.7  % Country211
      \\
    & & \no
      & 18.1  % ImageNet
      & 70.5  % CIFAR-10
      & 28.8  % CIFAR-100
      & 31.7  % SUN397
      & 44.8  % Caltech-101
      & 85.2  % STL-10
      & 9.4  % Food-101
      & 3.2  % CUB
      & 0.9  % Cars
      & 1.7  % Aircraft
      & \colorbox{winner}{11.0}  % Pets
      & \colorbox{winner}{8.7}  % Flowers-102
      & 10.7  % DTD
      & \colorbox{winner}{18.0}  % EuroSAT
      & 23.7  % RESISC45
      & 0.7  % Country211
      \\
    \cmidrule{2-19}
    & \multirow{2}{*}{ViT-B/16}
    & \yes
      & 19.6  % ImageNet
      & 77.5  % CIFAR-10
      & 34.2  % CIFAR-100
      & 34.6  % SUN397
      & \colorbox{winner}{48.4}  % Caltech-101
      & \colorbox{winner}{88.0}  % STL-10
      & 10.5  % Food-101
      & \colorbox{winner}{2.7}  % CUB
      & 1.2  % Cars
      & 0.5  % Aircraft
      & \colorbox{winner}{12.8}  % Pets
      & 8.6  % Flowers-102
      & 10.6  % DTD
      & \colorbox{winner}{17.6}  % EuroSAT
      & 22.0  % RESISC45
      & 0.8  % Country211
      \\
    & & \no
      & 19.6  % ImageNet
      & 77.5  % CIFAR-10
      & \colorbox{winner}{34.3}  % CIFAR-100
      & 34.6  % SUN397
      & 48.0  % Caltech-101
      & 87.9  % STL-10
      & 10.5  % Food-101
      & 2.6  % CUB
      & 1.2  % Cars
      & 0.5  % Aircraft
      & 12.7  % Pets
      & \colorbox{winner}{8.7}  % Flowers-102
      & \colorbox{winner}{10.7}  % DTD
      & 17.5  % EuroSAT
      & 22.0  % RESISC45
      & 0.8  % Country211
      \\
    \midrule
    \multirow{4}{*}{PHyCLIP}
    & \multirow{2}{*}{ViT-S/16}
    & \yes
      & 17.9  % ImageNet
      & \colorbox{winner}{68.0}  % CIFAR-10
      & \colorbox{winner}{27.6}  % CIFAR-100
      & \colorbox{winner}{31.8}  % SUN397
      & \colorbox{winner}{47.4}  % Caltech-101
      & 84.9  % STL-10
      & \colorbox{winner}{11.6}  % Food-101
      & 3.7  % CUB
      & 0.9  % Cars
      & 1.1  % Aircraft
      & 13.4  % Pets
      & 8.0  % Flowers-102
      & 10.5  % DTD
      & \colorbox{winner}{18.8}  % EuroSAT
      & \colorbox{winner}{22.2}  % RESISC45
      & 0.7  % Country211
      \\
    & & \no
      & 17.9  % ImageNet
      & 67.6  % CIFAR-10
      & 27.5  % CIFAR-100
      & 31.7  % SUN397
      & 47.3  % Caltech-101
      & \colorbox{winner}{85.0}  % STL-10
      & 11.3  % Food-101
      & \colorbox{winner}{3.8}  % CUB
      & \colorbox{winner}{1.0}  % Cars
      & \colorbox{winner}{1.3}  % Aircraft
      & \colorbox{winner}{13.6}  % Pets
      & \colorbox{winner}{8.2}  % Flowers-102
      & 10.5  % DTD
      & 18.5  % EuroSAT
      & 22.1  % RESISC45
      & 0.7  % Country211
      \\
    \cmidrule{2-19}
    & \multirow{2}{*}{ViT-B/16}
    & \yes
      & 20.1  % ImageNet
      & 74.0  % CIFAR-10
      & \colorbox{winner}{32.8}  % CIFAR-100
      & \colorbox{winner}{34.9}  % SUN397
      & \colorbox{winner}{52.7}  % Caltech-101
      & \colorbox{winner}{87.4}  % STL-10
      & \colorbox{winner}{13.5}  % Food-101
      & 3.6  % CUB
      & 1.5  % Cars
      & 1.3  % Aircraft
      & 11.9  % Pets
      & 9.5  % Flowers-102
      & 12.2  % DTD
      & \colorbox{winner}{20.6}  % EuroSAT
      & \colorbox{winner}{24.1}  % RESISC45
      & 0.8  % Country211
      \\
    & & \no
      & 20.1  % ImageNet
      & \colorbox{winner}{74.2}  % CIFAR-10
      & 32.4  % CIFAR-100
      & 34.8  % SUN397
      & 52.4  % Caltech-101
      & 87.3  % STL-10
      & 13.4  % Food-101
      & \colorbox{winner}{3.7}  % CUB
      & 1.5  % Cars
      & 1.3  % Aircraft
      & 11.9  % Pets
      & \colorbox{winner}{9.6}  % Flowers-102
      & \colorbox{winner}{12.4}  % DTD
      & 20.1  % EuroSAT
      & 23.7  % RESISC45
      & 0.8  % Country211
      \\
    \bottomrule
  \end{tabular}%
  }
\end{table}

\begin{table}[H]
  \centering
  \caption{Zero-shot image--text retrieval on COCO and Flickr30K, and hierarchical classification on WordNet over ImageNet, for each trained hyperbolic checkpoint read with the exponential map (\yes) and without it (\no). The \colorbox{winner}{shaded} cell marks the better reading in each column, per model and encoder; ties are unshaded.}
  \label{tab:expmap_inference_retrieval_hierarchical}
  \small
  \setlength{\tabcolsep}{4pt}
  \resizebox{\textwidth}{!}{%
  \begin{tabular}{@{}lllcccc cccc ccccc}
    \toprule
    & & &
    \multicolumn{4}{c}{\cellcolor{bandB}Text $\rightarrow$ Image} &
    \multicolumn{4}{c}{\cellcolor{bandB}Image $\rightarrow$ Text} &
    \multicolumn{5}{c}{\cellcolor{bandA}Hierarchical metrics} \\
    \cmidrule{4-16}
    & & &
    \multicolumn{2}{c}{\cellcolor{bandB}COCO} &
    \multicolumn{2}{c}{\cellcolor{bandB}Flickr30K} &
    \multicolumn{2}{c}{\cellcolor{bandB}COCO} &
    \multicolumn{2}{c}{\cellcolor{bandB}Flickr30K} &
    \multicolumn{5}{c}{\cellcolor{bandA}ImageNet} \\
    \cmidrule{4-16}
    Model & Encoder & $\expm$ &
    R@5 & R@10 & R@5 & R@10 & R@5 & R@10 & R@5 & R@10 &
    TIE($\downarrow$) & LCA($\downarrow$) & $J$($\uparrow$) & $P_{\mathrm{H}}$($\uparrow$) & $R_{\mathrm{H}}$($\uparrow$) \\
    \midrule
    \multirow{4}{*}{MERU}
    & \multirow{2}{*}{ViT-S/16}
    & \yes
      & 26.2  % T->I COCO R@5
      & 36.8  % T->I COCO R@10
      & 42.7  % T->I Flickr30K R@5
      & 55.1  % T->I Flickr30K R@10
      & 32.5  % I->T COCO R@5
      & 43.6  % I->T COCO R@10
      & 54.5  % I->T Flickr30K R@5
      & \colorbox{winner}{67.0}  % I->T Flickr30K R@10
      & \colorbox{winner}{6.582}  % ImageNet TIE (lower better)
      & 3.442  % ImageNet LCA (lower better)
      & 0.596  % ImageNet J (higher better)
      & 0.713  % ImageNet P_H (higher better)
      & 0.711  % ImageNet R_H (higher better)
      \\
    & & \no
      & 26.2  % T->I COCO R@5
      & 36.8  % T->I COCO R@10
      & 42.7  % T->I Flickr30K R@5
      & 55.1  % T->I Flickr30K R@10
      & \colorbox{winner}{32.7}  % I->T COCO R@5
      & \colorbox{winner}{43.7}  % I->T COCO R@10
      & 54.5  % I->T Flickr30K R@5
      & 66.9  % I->T Flickr30K R@10
      & 6.584  % ImageNet TIE (lower better)
      & 3.442  % ImageNet LCA (lower better)
      & 0.596  % ImageNet J (higher better)
      & 0.713  % ImageNet P_H (higher better)
      & 0.711  % ImageNet R_H (higher better)
      \\
    \cmidrule{2-16}
    & \multirow{2}{*}{ViT-B/16}
    & \yes
      & 29.8  % T->I COCO R@5
      & 40.8  % T->I COCO R@10
      & 48.1  % T->I Flickr30K R@5
      & 60.0  % T->I Flickr30K R@10
      & 36.4  % I->T COCO R@5
      & 48.8  % I->T COCO R@10
      & 58.8  % I->T Flickr30K R@5
      & 70.8  % I->T Flickr30K R@10
      & 6.220  % ImageNet TIE (lower better)
      & 3.304  % ImageNet LCA (lower better)
      & 0.621  % ImageNet J (higher better)
      & 0.730  % ImageNet P_H (higher better)
      & 0.732  % ImageNet R_H (higher better)
      \\
    & & \no
      & 29.8  % T->I COCO R@5
      & 40.8  % T->I COCO R@10
      & 48.1  % T->I Flickr30K R@5
      & 60.0  % T->I Flickr30K R@10
      & \colorbox{winner}{36.7}  % I->T COCO R@5
      & 48.8  % I->T COCO R@10
      & 58.8  % I->T Flickr30K R@5
      & 70.8  % I->T Flickr30K R@10
      & \colorbox{winner}{6.218}  % ImageNet TIE (lower better)
      & \colorbox{winner}{3.303}  % ImageNet LCA (lower better)
      & 0.621  % ImageNet J (higher better)
      & 0.730  % ImageNet P_H (higher better)
      & 0.732  % ImageNet R_H (higher better)
      \\
    \midrule
    \multirow{4}{*}{HyCoCLIP}
    & \multirow{2}{*}{ViT-S/16}
    & \yes
      & 30.1  % T->I COCO R@5
      & 41.1  % T->I COCO R@10
      & 49.1  % T->I Flickr30K R@5
      & 60.7  % T->I Flickr30K R@10
      & 37.1  % I->T COCO R@5
      & \colorbox{winner}{48.9}  % I->T COCO R@10
      & 61.3  % I->T Flickr30K R@5
      & 72.4  % I->T Flickr30K R@10
      & \colorbox{winner}{5.851}  % ImageNet TIE (lower better)
      & 3.095  % ImageNet LCA (lower better)
      & 0.641  % ImageNet J (higher better)
      & 0.748  % ImageNet P_H (higher better)
      & 0.747  % ImageNet R_H (higher better)
      \\
    & & \no
      & 30.1  % T->I COCO R@5
      & 41.1  % T->I COCO R@10
      & 49.1  % T->I Flickr30K R@5
      & 60.7  % T->I Flickr30K R@10
      & \colorbox{winner}{37.2}  % I->T COCO R@5
      & 48.8  % I->T COCO R@10
      & 61.3  % I->T Flickr30K R@5
      & \colorbox{winner}{72.5}  % I->T Flickr30K R@10
      & 5.852  % ImageNet TIE (lower better)
      & 3.095  % ImageNet LCA (lower better)
      & 0.641  % ImageNet J (higher better)
      & 0.748  % ImageNet P_H (higher better)
      & 0.747  % ImageNet R_H (higher better)
      \\
    \cmidrule{2-16}
    & \multirow{2}{*}{ViT-B/16}
    & \yes
      & 32.6  % T->I COCO R@5
      & 43.7  % T->I COCO R@10
      & 51.9  % T->I Flickr30K R@5
      & 63.1  % T->I Flickr30K R@10
      & \colorbox{winner}{39.4}  % I->T COCO R@5
      & 51.6  % I->T COCO R@10
      & 63.8  % I->T Flickr30K R@5
      & 73.7  % I->T Flickr30K R@10
      & 5.719  % ImageNet TIE (lower better)
      & 3.103  % ImageNet LCA (lower better)
      & 0.652  % ImageNet J (higher better)
      & 0.753  % ImageNet P_H (higher better)
      & 0.757  % ImageNet R_H (higher better)
      \\
    & & \no
      & 32.6  % T->I COCO R@5
      & 43.7  % T->I COCO R@10
      & 51.9  % T->I Flickr30K R@5
      & 63.1  % T->I Flickr30K R@10
      & 39.3  % I->T COCO R@5
      & 51.6  % I->T COCO R@10
      & \colorbox{winner}{64.1}  % I->T Flickr30K R@5
      & \colorbox{winner}{74.0}  % I->T Flickr30K R@10
      & \colorbox{winner}{5.718}  % ImageNet TIE (lower better)
      & 3.103  % ImageNet LCA (lower better)
      & \colorbox{winner}{0.653}  % ImageNet J (higher better)
      & 0.753  % ImageNet P_H (higher better)
      & 0.757  % ImageNet R_H (higher better)
      \\
    \midrule
    \multirow{4}{*}{ARGENT}
    & \multirow{2}{*}{ViT-S/16}
    & \yes
      & 30.1  % T->I COCO R@5
      & 41.2  % T->I COCO R@10
      & 46.7  % T->I Flickr30K R@5
      & 58.1  % T->I Flickr30K R@10
      & \colorbox{winner}{35.8}  % I->T COCO R@5
      & 47.7  % I->T COCO R@10
      & \colorbox{winner}{57.6}  % I->T Flickr30K R@5
      & \colorbox{winner}{69.5}  % I->T Flickr30K R@10
      & 5.931  % ImageNet TIE (lower better)
      & 3.132  % ImageNet LCA (lower better)
      & 0.637  % ImageNet J (higher better)
      & 0.746  % ImageNet P_H (higher better)
      & 0.742  % ImageNet R_H (higher better)
      \\
    & & \no
      & 30.1  % T->I COCO R@5
      & 41.2  % T->I COCO R@10
      & \colorbox{winner}{46.8}  % T->I Flickr30K R@5
      & 58.1  % T->I Flickr30K R@10
      & 35.7  % I->T COCO R@5
      & \colorbox{winner}{47.9}  % I->T COCO R@10
      & 56.9  % I->T Flickr30K R@5
      & 69.3  % I->T Flickr30K R@10
      & \colorbox{winner}{5.926}  % ImageNet TIE (lower better)
      & \colorbox{winner}{3.127}  % ImageNet LCA (lower better)
      & 0.637  % ImageNet J (higher better)
      & 0.746  % ImageNet P_H (higher better)
      & 0.742  % ImageNet R_H (higher better)
      \\
    \cmidrule{2-16}
    & \multirow{2}{*}{ViT-B/16}
    & \yes
      & 32.5  % T->I COCO R@5
      & 44.0  % T->I COCO R@10
      & 52.4  % T->I Flickr30K R@5
      & 63.2  % T->I Flickr30K R@10
      & \colorbox{winner}{39.5}  % I->T COCO R@5
      & 51.7  % I->T COCO R@10
      & 62.9  % I->T Flickr30K R@5
      & 74.3  % I->T Flickr30K R@10
      & 5.661  % ImageNet TIE (lower better)
      & 3.079  % ImageNet LCA (lower better)
      & 0.656  % ImageNet J (higher better)
      & 0.757  % ImageNet P_H (higher better)
      & 0.761  % ImageNet R_H (higher better)
      \\
    & & \no
      & 32.5  % T->I COCO R@5
      & 44.0  % T->I COCO R@10
      & 52.4  % T->I Flickr30K R@5
      & 63.2  % T->I Flickr30K R@10
      & 39.4  % I->T COCO R@5
      & 51.7  % I->T COCO R@10
      & 62.9  % I->T Flickr30K R@5
      & \colorbox{winner}{74.4}  % I->T Flickr30K R@10
      & \colorbox{winner}{5.658}  % ImageNet TIE (lower better)
      & \colorbox{winner}{3.077}  % ImageNet LCA (lower better)
      & 0.656  % ImageNet J (higher better)
      & 0.757  % ImageNet P_H (higher better)
      & 0.761  % ImageNet R_H (higher better)
      \\
    \midrule
    \multirow{4}{*}{PHyCLIP}
    & \multirow{2}{*}{ViT-S/16}
    & \yes
      & 30.3  % T->I COCO R@5
      & 41.1  % T->I COCO R@10
      & \colorbox{winner}{48.3}  % T->I Flickr30K R@5
      & 59.2  % T->I Flickr30K R@10
      & 37.5  % I->T COCO R@5
      & 49.2  % I->T COCO R@10
      & \colorbox{winner}{58.4}  % I->T Flickr30K R@5
      & \colorbox{winner}{70.6}  % I->T Flickr30K R@10
      & 5.998  % ImageNet TIE (lower better)
      & 3.203  % ImageNet LCA (lower better)
      & 0.633  % ImageNet J (higher better)
      & 0.740  % ImageNet P_H (higher better)
      & 0.742  % ImageNet R_H (higher better)
      \\
    & & \no
      & \colorbox{winner}{30.5}  % T->I COCO R@5
      & \colorbox{winner}{41.2}  % T->I COCO R@10
      & 48.1  % T->I Flickr30K R@5
      & \colorbox{winner}{59.4}  % T->I Flickr30K R@10
      & \colorbox{winner}{37.6}  % I->T COCO R@5
      & \colorbox{winner}{49.3}  % I->T COCO R@10
      & 57.7  % I->T Flickr30K R@5
      & 70.1  % I->T Flickr30K R@10
      & \colorbox{winner}{5.995}  % ImageNet TIE (lower better)
      & \colorbox{winner}{3.198}  % ImageNet LCA (lower better)
      & 0.633  % ImageNet J (higher better)
      & \colorbox{winner}{0.741}  % ImageNet P_H (higher better)
      & 0.742  % ImageNet R_H (higher better)
      \\
    \cmidrule{2-16}
    & \multirow{2}{*}{ViT-B/16}
    & \yes
      & 32.8  % T->I COCO R@5
      & 44.0  % T->I COCO R@10
      & 51.8  % T->I Flickr30K R@5
      & 62.1  % T->I Flickr30K R@10
      & 39.6  % I->T COCO R@5
      & 52.1  % I->T COCO R@10
      & 62.8  % I->T Flickr30K R@5
      & 73.6  % I->T Flickr30K R@10
      & \colorbox{winner}{5.745}  % ImageNet TIE (lower better)
      & \colorbox{winner}{3.122}  % ImageNet LCA (lower better)
      & 0.652  % ImageNet J (higher better)
      & 0.752  % ImageNet P_H (higher better)
      & 0.755  % ImageNet R_H (higher better)
      \\
    & & \no
      & \colorbox{winner}{33.0}  % T->I COCO R@5
      & \colorbox{winner}{44.2}  % T->I COCO R@10
      & \colorbox{winner}{51.9}  % T->I Flickr30K R@5
      & \colorbox{winner}{62.4}  % T->I Flickr30K R@10
      & \colorbox{winner}{40.0}  % I->T COCO R@5
      & \colorbox{winner}{52.6}  % I->T COCO R@10
      & 62.8  % I->T Flickr30K R@5
      & \colorbox{winner}{73.8}  % I->T Flickr30K R@10
      & 5.747  % ImageNet TIE (lower better)
      & 3.125  % ImageNet LCA (lower better)
      & 0.652  % ImageNet J (higher better)
      & 0.752  % ImageNet P_H (higher better)
      & \colorbox{winner}{0.756}  % ImageNet R_H (higher better)
      \\
    \bottomrule
  \end{tabular}%
  }
\end{table}

\label{app:released_checkpoints}

\begin{table}[H]
  \centering
  \caption{Zero-shot classification accuracy (\%) across the sixteen benchmarks, for the publicly released ViT-B/16 checkpoints read with the exponential map (\yes) and without it (\no). The \colorbox{winner}{shaded} cell marks the better reading in each column, per model; ties are unshaded.}
  \label{tab:released_classification}
  \small
  \setlength{\tabcolsep}{8pt}
  \resizebox{\textwidth}{!}{%
  \begin{tabular}{@{}llcccccc cccccc cccc}
    \toprule
    & &
    \multicolumn{6}{c}{\cellcolor{bandB}General} &
    \multicolumn{6}{c}{\cellcolor{bandA}Fine-grained} &
    \multicolumn{4}{c}{\cellcolor{bandC}Specialized} \\
    \cmidrule{3-18}
    Model & $\expm$ &
    \cellcolor{bandB!80}\rotatebox{90}{ImageNet} &
    \cellcolor{bandB!80}\rotatebox{90}{CIFAR-10} &
    \cellcolor{bandB!80}\rotatebox{90}{CIFAR-100} &
    \cellcolor{bandB!80}\rotatebox{90}{SUN397} &
    \cellcolor{bandB!80}\rotatebox{90}{Caltech-101} &
    \cellcolor{bandB!80}\rotatebox{90}{STL-10} &
    \cellcolor{bandA!80}\rotatebox{90}{Food-101} &
    \cellcolor{bandA!80}\rotatebox{90}{CUB} &
    \cellcolor{bandA!80}\rotatebox{90}{Cars} &
    \cellcolor{bandA!80}\rotatebox{90}{Aircraft} &
    \cellcolor{bandA!80}\rotatebox{90}{Pets} &
    \cellcolor{bandA!80}\rotatebox{90}{Flowers-102} &
    \cellcolor{bandC!80}\rotatebox{90}{DTD} &
    \cellcolor{bandC!80}\rotatebox{90}{EuroSAT} &
    \cellcolor{bandC!80}\rotatebox{90}{RESISC45} &
    \cellcolor{bandC!80}\rotatebox{90}{Country211} \\
    \midrule
    \multirow{2}{*}{MERU}
    & \yes
      & 37.5  % ImageNet
      & 67.7  % CIFAR-10
      & \colorbox{winner}{32.7}  % CIFAR-100
      & 30.9  % SUN397
      & 68.1  % Caltech-101
      & 92.5  % STL-10
      & 78.7  % Food-101
      & 34.8  % CUB
      & 13.8  % Cars
      & 1.7  % Aircraft
      & 79.2  % Pets
      & 52.1  % Flowers-102
      & \colorbox{winner}{17.4}  % DTD
      & 30.2  % EuroSAT
      & 34.5  % RESISC45
      & 5.6  % Country211
      \\
    & \no
      & 37.5  % ImageNet
      & 67.7  % CIFAR-10
      & 32.6  % CIFAR-100
      & 30.9  % SUN397
      & 68.1  % Caltech-101
      & 92.5  % STL-10
      & 78.7  % Food-101
      & 34.8  % CUB
      & \colorbox{winner}{13.9}  % Cars
      & \colorbox{winner}{1.8}  % Aircraft
      & \colorbox{winner}{79.3}  % Pets
      & 52.1  % Flowers-102
      & 17.3  % DTD
      & 30.2  % EuroSAT
      & 34.5  % RESISC45
      & \colorbox{winner}{5.7}  % Country211
      \\
    \midrule
    \multirow{2}{*}{HyCoCLIP}
    & \yes
      & 45.8  % ImageNet
      & 88.8  % CIFAR-10
      & 60.1  % CIFAR-100
      & 57.2  % SUN397
      & 80.8  % Caltech-101
      & 95.0  % STL-10
      & 59.2  % Food-101
      & 16.4  % CUB
      & 11.6  % Cars
      & 3.7  % Aircraft
      & 56.8  % Pets
      & 23.9  % Flowers-102
      & \colorbox{winner}{29.4}  % DTD
      & 35.8  % EuroSAT
      & 45.6  % RESISC45
      & 6.5  % Country211
      \\
    & \no
      & 45.8  % ImageNet
      & 88.8  % CIFAR-10
      & \colorbox{winner}{60.2}  % CIFAR-100
      & 57.2  % SUN397
      & \colorbox{winner}{80.9}  % Caltech-101
      & 95.0  % STL-10
      & 59.2  % Food-101
      & 16.4  % CUB
      & 11.6  % Cars
      & 3.7  % Aircraft
      & \colorbox{winner}{56.9}  % Pets
      & 23.9  % Flowers-102
      & 29.2  % DTD
      & 35.8  % EuroSAT
      & 45.6  % RESISC45
      & 6.5  % Country211
      \\
    \bottomrule
  \end{tabular}}
\end{table}

\begin{table}[H]
  \centering
  \caption{Zero-shot image--text retrieval on COCO and Flickr30K, and hierarchical classification on WordNet over ImageNet, for the publicly released ViT-B/16 checkpoints read with the exponential map (\yes) and without it (\no). The \colorbox{winner}{shaded} cell marks the better reading in each column, per model; ties are unshaded.}
  \label{tab:released_retrieval_hierarchical}
  \small
  \setlength{\tabcolsep}{4pt}
  \resizebox{\textwidth}{!}{%
  \begin{tabular}{@{}llcccc cccc ccccc}
    \toprule
    & &
    \multicolumn{4}{c}{\cellcolor{bandB}Text $\rightarrow$ Image} &
    \multicolumn{4}{c}{\cellcolor{bandB}Image $\rightarrow$ Text} &
    \multicolumn{5}{c}{\cellcolor{bandA}Hierarchical metrics} \\
    \cmidrule{3-15}
    & &
    \multicolumn{2}{c}{\cellcolor{bandB}COCO} &
    \multicolumn{2}{c}{\cellcolor{bandB}Flickr30K} &
    \multicolumn{2}{c}{\cellcolor{bandB}COCO} &
    \multicolumn{2}{c}{\cellcolor{bandB}Flickr30K} &
    \multicolumn{5}{c}{\cellcolor{bandA}ImageNet} \\
    \cmidrule{3-15}
    Model & $\expm$ &
    R@5 & R@10 & R@5 & R@10 & R@5 & R@10 & R@5 & R@10 &
    TIE($\downarrow$) & LCA($\downarrow$) & $J$($\uparrow$) & $P_{\mathrm{H}}$($\uparrow$) & $R_{\mathrm{H}}$($\uparrow$) \\
    \midrule
    \multirow{2}{*}{MERU}
    & \yes
      & 33.2  % T->I COCO R@5
      & 44.0  % T->I COCO R@10
      & 41.1  % T->I Flickr30K R@5
      & 51.6  % T->I Flickr30K R@10
      & 41.8  % I->T COCO R@5
      & 52.9  % I->T COCO R@10
      & 48.1  % I->T Flickr30K R@5
      & 58.9  % I->T Flickr30K R@10
      & 4.033  % ImageNet TIE (lower better)
      & 2.418  % ImageNet LCA (lower better)
      & 0.756  % ImageNet J (higher better)
      & 0.829  % ImageNet P_H (higher better)
      & 0.832  % ImageNet R_H (higher better)
      \\
    & \no
      & 33.2  % T->I COCO R@5
      & 44.0  % T->I COCO R@10
      & 41.1  % T->I Flickr30K R@5
      & 51.6  % T->I Flickr30K R@10
      & \colorbox{winner}{41.9}  % I->T COCO R@5
      & \colorbox{winner}{53.1}  % I->T COCO R@10
      & \colorbox{winner}{48.6}  % I->T Flickr30K R@5
      & \colorbox{winner}{59.4}  % I->T Flickr30K R@10
      & \colorbox{winner}{4.028}  % ImageNet TIE (lower better)
      & \colorbox{winner}{2.417}  % ImageNet LCA (lower better)
      & 0.756  % ImageNet J (higher better)
      & 0.829  % ImageNet P_H (higher better)
      & 0.832  % ImageNet R_H (higher better)
      \\
    \midrule
    \multirow{2}{*}{HyCoCLIP}
    & \yes
      & 58.4  % T->I COCO R@5
      & 69.3  % T->I COCO R@10
      & 84.9  % T->I Flickr30K R@5
      & 90.3  % T->I Flickr30K R@10
      & \colorbox{winner}{72.0}  % I->T COCO R@5
      & 82.0  % I->T COCO R@10
      & 92.6  % I->T Flickr30K R@5
      & 95.4  % I->T Flickr30K R@10
      & \colorbox{winner}{3.172}  % ImageNet TIE (lower better)
      & \colorbox{winner}{2.047}  % ImageNet LCA (lower better)
      & 0.814  % ImageNet J (higher better)
      & \colorbox{winner}{0.874}  % ImageNet P_H (higher better)
      & 0.874  % ImageNet R_H (higher better)
      \\
    & \no
      & 58.4  % T->I COCO R@5
      & 69.3  % T->I COCO R@10
      & 84.9  % T->I Flickr30K R@5
      & 90.3  % T->I Flickr30K R@10
      & 71.9  % I->T COCO R@5
      & 82.0  % I->T COCO R@10
      & 92.6  % I->T Flickr30K R@5
      & 95.4  % I->T Flickr30K R@10
      & 3.175  % ImageNet TIE (lower better)
      & 2.048  % ImageNet LCA (lower better)
      & 0.814  % ImageNet J (higher better)
      & 0.873  % ImageNet P_H (higher better)
      & 0.874  % ImageNet R_H (higher better)
      \\
    \bottomrule
  \end{tabular}}
\end{table}

\begin{figure}[H]
  \centering
  \includegraphics[width=\textwidth]{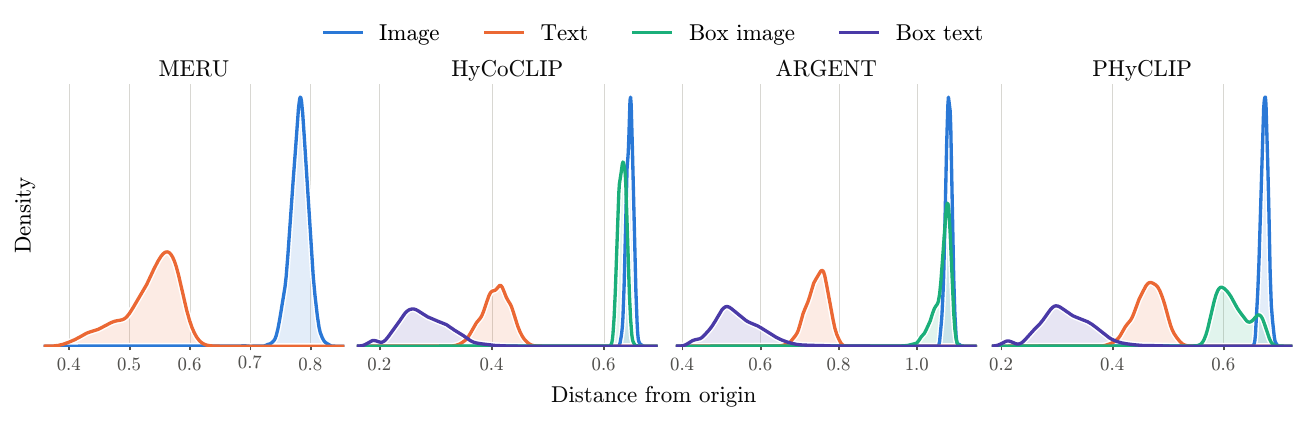}\vspace{-1cm}
  \caption{Radial distribution of the embeddings for the Euclidean controls, in the same layout as \cref{fig:radial_distribution}: the ViT-S/16 models on CC3M, one kernel density per modality over distance from the origin.}
  \label{fig:radial_distribution_euclidean}
\end{figure}\vspace{-0.2cm}

\begin{table}[H]
  \centering
  \caption{Zero-shot classification accuracy (\%) across the sixteen benchmarks, for each objective in its hyperbolic formulation ($\mathbb{L}^{n}_{c}$) against its Euclidean control ($\mathbb{E}^{n}$). The \colorbox{winner}{shaded} cell marks the better geometry in each column, per encoder; ties are unshaded.}
  \label{tab:euclidean_training_classification}
  \small
  \setlength{\tabcolsep}{8pt}
  \resizebox{\textwidth}{!}{%
  \begin{tabular}{@{}lllcccccc cccccc cccc}
    \toprule
    & & &
    \multicolumn{6}{c}{\cellcolor{bandB}General} &
    \multicolumn{6}{c}{\cellcolor{bandA}Fine-grained} &
    \multicolumn{4}{c}{\cellcolor{bandC}Specialized} \\
    \cmidrule{4-19}
    Model & Encoder & Geometry &
    \cellcolor{bandB!80}\rotatebox{90}{ImageNet} &
    \cellcolor{bandB!80}\rotatebox{90}{CIFAR-10} &
    \cellcolor{bandB!80}\rotatebox{90}{CIFAR-100} &
    \cellcolor{bandB!80}\rotatebox{90}{SUN397} &
    \cellcolor{bandB!80}\rotatebox{90}{Caltech-101} &
    \cellcolor{bandB!80}\rotatebox{90}{STL-10} &
    \cellcolor{bandA!80}\rotatebox{90}{Food-101} &
    \cellcolor{bandA!80}\rotatebox{90}{CUB} &
    \cellcolor{bandA!80}\rotatebox{90}{Cars} &
    \cellcolor{bandA!80}\rotatebox{90}{Aircraft} &
    \cellcolor{bandA!80}\rotatebox{90}{Pets} &
    \cellcolor{bandA!80}\rotatebox{90}{Flowers-102} &
    \cellcolor{bandC!80}\rotatebox{90}{DTD} &
    \cellcolor{bandC!80}\rotatebox{90}{EuroSAT} &
    \cellcolor{bandC!80}\rotatebox{90}{RESISC45} &
    \cellcolor{bandC!80}\rotatebox{90}{Country211} \\
    \midrule
    \multirow{4}{*}{MERU}
    & \multirow{2}{*}{ViT-S/16}
    & $\mathbb{E}^{n}$
      & \best{15.0}  % ImageNet
      & 47.1  % CIFAR-10
      & 20.0  % CIFAR-100
      & \best{28.5}  % SUN397
      & 41.5  % Caltech-101
      & \best{78.6}  % STL-10
      & 8.1  % Food-101
      & \best{3.2}  % CUB
      & \best{1.1}  % Cars
      & 1.0  % Aircraft
      & \best{11.6}  % Pets
      & \best{8.2}  % Flowers-102
      & \best{9.4}  % DTD
      & \best{20.3}  % EuroSAT
      & \best{23.0}  % RESISC45
      & 0.6  % Country211
      \\
    & & $\mathbb{L}^{n}_{c}$
      & 14.3  % ImageNet
      & \best{53.1}  % CIFAR-10
      & \best{22.9}  % CIFAR-100
      & 27.6  % SUN397
      & \best{42.1}  % Caltech-101
      & 75.3  % STL-10
      & \best{9.3}  % Food-101
      & 3.0  % CUB
      & 0.9  % Cars
      & \best{1.2}  % Aircraft
      & 9.8  % Pets
      & 7.3  % Flowers-102
      & 8.0  % DTD
      & 18.4  % EuroSAT
      & 20.4  % RESISC45
      & \best{0.8}  % Country211
      \\
    \cmidrule{2-19}
    & \multirow{2}{*}{ViT-B/16}
    & $\mathbb{E}^{n}$
      & \best{17.2}  % ImageNet
      & \best{60.8}  % CIFAR-10
      & \best{26.4}  % CIFAR-100
      & \best{31.3}  % SUN397
      & 46.9  % Caltech-101
      & 83.4  % STL-10
      & \best{11.1}  % Food-101
      & \best{3.2}  % CUB
      & \best{1.4}  % Cars
      & 1.3  % Aircraft
      & \best{13.6}  % Pets
      & 8.3  % Flowers-102
      & 10.1  % DTD
      & \best{20.1}  % EuroSAT
      & \best{24.8}  % RESISC45
      & \best{0.8}  % Country211
      \\
    & & $\mathbb{L}^{n}_{c}$
      & 16.7  % ImageNet
      & 60.5  % CIFAR-10
      & 25.9  % CIFAR-100
      & 30.9  % SUN397
      & \best{47.5}  % Caltech-101
      & \best{84.0}  % STL-10
      & 10.7  % Food-101
      & 2.7  % CUB
      & 1.2  % Cars
      & \best{1.7}  % Aircraft
      & 10.9  % Pets
      & \best{9.0}  % Flowers-102
      & \best{10.8}  % DTD
      & 19.1  % EuroSAT
      & 23.1  % RESISC45
      & 0.7  % Country211
      \\
    \midrule
    \multirow{4}{*}{HyCoCLIP}
    & \multirow{2}{*}{ViT-S/16}
    & $\mathbb{E}^{n}$
      & 17.9  % ImageNet
      & 66.1  % CIFAR-10
      & 29.3  % CIFAR-100
      & 30.8  % SUN397
      & 47.3  % Caltech-101
      & 83.0  % STL-10
      & \best{10.9}  % Food-101
      & 3.3  % CUB
      & \best{1.3}  % Cars
      & 1.5  % Aircraft
      & 12.1  % Pets
      & \best{8.6}  % Flowers-102
      & 12.2  % DTD
      & \best{28.8}  % EuroSAT
      & 22.2  % RESISC45
      & 0.7  % Country211
      \\
    & & $\mathbb{L}^{n}_{c}$
      & \best{18.4}  % ImageNet
      & \best{72.6}  % CIFAR-10
      & \best{31.3}  % CIFAR-100
      & \best{33.0}  % SUN397
      & \best{50.0}  % Caltech-101
      & \best{85.0}  % STL-10
      & 10.4  % Food-101
      & \best{3.6}  % CUB
      & 1.1  % Cars
      & \best{1.6}  % Aircraft
      & \best{15.1}  % Pets
      & 7.8  % Flowers-102
      & \best{13.5}  % DTD
      & 13.1  % EuroSAT
      & \best{23.6}  % RESISC45
      & 0.7  % Country211
      \\
    \cmidrule{2-19}
    & \multirow{2}{*}{ViT-B/16}
    & $\mathbb{E}^{n}$
      & \best{21.2}  % ImageNet
      & \best{76.2}  % CIFAR-10
      & \best{35.6}  % CIFAR-100
      & \best{34.9}  % SUN397
      & \best{52.3}  % Caltech-101
      & \best{89.5}  % STL-10
      & \best{14.5}  % Food-101
      & \best{4.0}  % CUB
      & \best{1.4}  % Cars
      & 1.2  % Aircraft
      & 13.4  % Pets
      & \best{10.4}  % Flowers-102
      & \best{12.7}  % DTD
      & \best{21.6}  % EuroSAT
      & \best{25.4}  % RESISC45
      & 0.6  % Country211
      \\
    & & $\mathbb{L}^{n}_{c}$
      & 19.9  % ImageNet
      & 75.0  % CIFAR-10
      & 34.1  % CIFAR-100
      & 34.8  % SUN397
      & 50.3  % Caltech-101
      & 88.5  % STL-10
      & 11.7  % Food-101
      & 3.8  % CUB
      & 1.3  % Cars
      & 1.2  % Aircraft
      & \best{16.2}  % Pets
      & 7.6  % Flowers-102
      & 12.0  % DTD
      & 19.1  % EuroSAT
      & 24.0  % RESISC45
      & \best{0.7}  % Country211
      \\
    \midrule
    \multirow{4}{*}{ARGENT}
    & \multirow{2}{*}{ViT-S/16}
    & $\mathbb{E}^{n}$
      & 17.6  % ImageNet
      & 69.4  % CIFAR-10
      & 28.1  % CIFAR-100
      & \best{31.9}  % SUN397
      & 44.2  % Caltech-101
      & \best{86.0}  % STL-10
      & 9.0  % Food-101
      & 3.2  % CUB
      & \best{1.2}  % Cars
      & 1.2  % Aircraft
      & \best{13.8}  % Pets
      & 8.6  % Flowers-102
      & 10.1  % DTD
      & 16.1  % EuroSAT
      & 21.9  % RESISC45
      & \best{0.8}  % Country211
      \\
    & & $\mathbb{L}^{n}_{c}$
      & \best{18.1}  % ImageNet
      & \best{70.7}  % CIFAR-10
      & \best{28.9}  % CIFAR-100
      & 31.8  % SUN397
      & \best{45.2}  % Caltech-101
      & 85.2  % STL-10
      & \best{9.5}  % Food-101
      & 3.2  % CUB
      & 0.9  % Cars
      & \best{1.8}  % Aircraft
      & 10.9  % Pets
      & 8.6  % Flowers-102
      & \best{10.9}  % DTD
      & \best{17.9}  % EuroSAT
      & \best{23.8}  % RESISC45
      & 0.7  % Country211
      \\
    \cmidrule{2-19}
    & \multirow{2}{*}{ViT-B/16}
    & $\mathbb{E}^{n}$
      & \best{20.7}  % ImageNet
      & \best{77.8}  % CIFAR-10
      & 33.8  % CIFAR-100
      & 34.1  % SUN397
      & \best{51.9}  % Caltech-101
      & 87.8  % STL-10
      & \best{11.0}  % Food-101
      & \best{4.1}  % CUB
      & \best{1.3}  % Cars
      & \best{1.1}  % Aircraft
      & \best{14.3}  % Pets
      & \best{11.1}  % Flowers-102
      & \best{12.9}  % DTD
      & \best{22.2}  % EuroSAT
      & \best{24.2}  % RESISC45
      & 0.7  % Country211
      \\
    & & $\mathbb{L}^{n}_{c}$
      & 19.6  % ImageNet
      & 77.5  % CIFAR-10
      & \best{34.2}  % CIFAR-100
      & \best{34.6}  % SUN397
      & 48.4  % Caltech-101
      & \best{88.0}  % STL-10
      & 10.5  % Food-101
      & 2.7  % CUB
      & 1.2  % Cars
      & 0.5  % Aircraft
      & 12.8  % Pets
      & 8.6  % Flowers-102
      & 10.6  % DTD
      & 17.6  % EuroSAT
      & 22.0  % RESISC45
      & \best{0.8}  % Country211
      \\
    \midrule
    \multirow{4}{*}{PHyCLIP}
    & \multirow{2}{*}{ViT-S/16}
    & $\prod_{i=1}^{k}\mathbb{E}^{d}$
      & \best{18.0}  % ImageNet
      & \best{68.5}  % CIFAR-10
      & \best{28.4}  % CIFAR-100
      & \best{31.9}  % SUN397
      & \best{49.3}  % Caltech-101
      & 83.9  % STL-10
      & \best{11.8}  % Food-101
      & 3.7  % CUB
      & 0.9  % Cars
      & 1.1  % Aircraft
      & 11.4  % Pets
      & \best{8.4}  % Flowers-102
      & 8.0  % DTD
      & \best{27.9}  % EuroSAT
      & \best{22.9}  % RESISC45
      & 0.7  % Country211
      \\
    & & $\prod_{i=1}^{k}\mathbb{L}^{d}_{c_i}$
      & 17.9  % ImageNet
      & 68.0  % CIFAR-10
      & 27.6  % CIFAR-100
      & 31.8  % SUN397
      & 47.4  % Caltech-101
      & \best{84.9}  % STL-10
      & 11.6  % Food-101
      & 3.7  % CUB
      & 0.9  % Cars
      & 1.1  % Aircraft
      & \best{13.4}  % Pets
      & 8.0  % Flowers-102
      & \best{10.5}  % DTD
      & 18.8  % EuroSAT
      & 22.2  % RESISC45
      & 0.7  % Country211
      \\
    \cmidrule{2-19}
    & \multirow{2}{*}{ViT-B/16}
    & $\prod_{i=1}^{k}\mathbb{E}^{d}$
      & \best{20.5}  % ImageNet
      & \best{75.1}  % CIFAR-10
      & \best{33.5}  % CIFAR-100
      & 34.8  % SUN397
      & \best{52.8}  % Caltech-101
      & \best{89.4}  % STL-10
      & 13.1  % Food-101
      & \best{5.1}  % CUB
      & 1.2  % Cars
      & \best{1.7}  % Aircraft
      & \best{12.7}  % Pets
      & 8.5  % Flowers-102
      & \best{13.9}  % DTD
      & \best{21.2}  % EuroSAT
      & 23.2  % RESISC45
      & 0.8  % Country211
      \\
    & & $\prod_{i=1}^{k}\mathbb{L}^{d}_{c_i}$
      & 20.1  % ImageNet
      & 74.0  % CIFAR-10
      & 32.8  % CIFAR-100
      & \best{34.9}  % SUN397
      & 52.7  % Caltech-101
      & 87.4  % STL-10
      & \best{13.5}  % Food-101
      & 3.6  % CUB
      & \best{1.5}  % Cars
      & 1.3  % Aircraft
      & 11.9  % Pets
      & \best{9.5}  % Flowers-102
      & 12.2  % DTD
      & 20.6  % EuroSAT
      & \best{24.1}  % RESISC45
      & 0.8  % Country211
      \\
    \bottomrule
  \end{tabular}%
  }
\end{table}

\begin{table}[H]
  \centering
  \caption{Zero-shot image--text retrieval on COCO and Flickr30K, and hierarchical classification on WordNet over ImageNet, for each objective against its Euclidean control. The \colorbox{winner}{shaded} cell marks the better geometry in each column, per encoder; ties are unshaded.}
  \label{tab:euclidean_training_retrieval_hierarchical}
  \small
  \setlength{\tabcolsep}{4pt}
  \resizebox{\textwidth}{!}{%
  \begin{tabular}{@{}lllcccc cccc ccccc}
    \toprule
    & & &
    \multicolumn{4}{c}{\cellcolor{bandB}Text $\rightarrow$ Image} &
    \multicolumn{4}{c}{\cellcolor{bandB}Image $\rightarrow$ Text} &
    \multicolumn{5}{c}{\cellcolor{bandA}Hierarchical metrics} \\
    \cmidrule{4-16}
    & & &
    \multicolumn{2}{c}{\cellcolor{bandB}COCO} &
    \multicolumn{2}{c}{\cellcolor{bandB}Flickr30K} &
    \multicolumn{2}{c}{\cellcolor{bandB}COCO} &
    \multicolumn{2}{c}{\cellcolor{bandB}Flickr30K} &
    \multicolumn{5}{c}{\cellcolor{bandA}ImageNet} \\
    \cmidrule{4-16}
    Model & Encoder & Geometry &
    R@5 & R@10 & R@5 & R@10 & R@5 & R@10 & R@5 & R@10 &
    TIE($\downarrow$) & LCA($\downarrow$) & $J$($\uparrow$) & $P_{\mathrm{H}}$($\uparrow$) & $R_{\mathrm{H}}$($\uparrow$) \\
    \midrule
    \multirow{4}{*}{MERU}
    & \multirow{2}{*}{ViT-S/16}
    & $\mathbb{E}^{n}$
      & \best{26.7}  % T->I COCO R@5
      & \best{36.9}  % T->I COCO R@10
      & \best{44.1}  % T->I Flickr30K R@5
      & \best{55.6}  % T->I Flickr30K R@10
      & \best{33.3}  % I->T COCO R@5
      & \best{45.0}  % I->T COCO R@10
      & \best{56.6}  % I->T Flickr30K R@5
      & \best{67.2}  % I->T Flickr30K R@10
      & \best{6.440}  % ImageNet TIE (lower better)
      & \best{3.384}  % ImageNet LCA (lower better)
      & \best{0.605}  % ImageNet J (higher better)
      & \best{0.719}  % ImageNet P_H (higher better)
      & \best{0.719}  % ImageNet R_H (higher better)
      \\
    & & $\mathbb{L}^{n}_{c}$
      & 26.2  % T->I COCO R@5
      & 36.8  % T->I COCO R@10
      & 42.7  % T->I Flickr30K R@5
      & 55.1  % T->I Flickr30K R@10
      & 32.5  % I->T COCO R@5
      & 43.6  % I->T COCO R@10
      & 54.5  % I->T Flickr30K R@5
      & 67.0  % I->T Flickr30K R@10
      & 6.582  % ImageNet TIE (lower better)
      & 3.442  % ImageNet LCA (lower better)
      & 0.596  % ImageNet J (higher better)
      & 0.713  % ImageNet P_H (higher better)
      & 0.711  % ImageNet R_H (higher better)
      \\
    \cmidrule{2-16}
    & \multirow{2}{*}{ViT-B/16}
    & $\mathbb{E}^{n}$
      & 29.7  % T->I COCO R@5
      & 40.6  % T->I COCO R@10
      & \best{48.6}  % T->I Flickr30K R@5
      & 59.4  % T->I Flickr30K R@10
      & \best{37.5}  % I->T COCO R@5
      & \best{49.2}  % I->T COCO R@10
      & \best{61.5}  % I->T Flickr30K R@5
      & \best{73.2}  % I->T Flickr30K R@10
      & \best{6.091}  % ImageNet TIE (lower better)
      & \best{3.215}  % ImageNet LCA (lower better)
      & \best{0.626}  % ImageNet J (higher better)
      & \best{0.737}  % ImageNet P_H (higher better)
      & \best{0.735}  % ImageNet R_H (higher better)
      \\
    & & $\mathbb{L}^{n}_{c}$
      & \best{29.8}  % T->I COCO R@5
      & \best{40.8}  % T->I COCO R@10
      & 48.1  % T->I Flickr30K R@5
      & \best{60.0}  % T->I Flickr30K R@10
      & 36.4  % I->T COCO R@5
      & 48.8  % I->T COCO R@10
      & 58.8  % I->T Flickr30K R@5
      & 70.8  % I->T Flickr30K R@10
      & 6.220  % ImageNet TIE (lower better)
      & 3.304  % ImageNet LCA (lower better)
      & 0.621  % ImageNet J (higher better)
      & 0.730  % ImageNet P_H (higher better)
      & 0.732  % ImageNet R_H (higher better)
      \\
    \midrule
    \multirow{4}{*}{HyCoCLIP}
    & \multirow{2}{*}{ViT-S/16}
    & $\mathbb{E}^{n}$
      & 28.3  % T->I COCO R@5
      & 39.1  % T->I COCO R@10
      & 45.9  % T->I Flickr30K R@5
      & 56.9  % T->I Flickr30K R@10
      & 35.1  % I->T COCO R@5
      & 46.5  % I->T COCO R@10
      & 55.6  % I->T Flickr30K R@5
      & 68.5  % I->T Flickr30K R@10
      & 6.028  % ImageNet TIE (lower better)
      & 3.180  % ImageNet LCA (lower better)
      & 0.632  % ImageNet J (higher better)
      & 0.740  % ImageNet P_H (higher better)
      & 0.738  % ImageNet R_H (higher better)
      \\
    & & $\mathbb{L}^{n}_{c}$
      & \best{30.1}  % T->I COCO R@5
      & \best{41.1}  % T->I COCO R@10
      & \best{49.1}  % T->I Flickr30K R@5
      & \best{60.7}  % T->I Flickr30K R@10
      & \best{37.1}  % I->T COCO R@5
      & \best{48.9}  % I->T COCO R@10
      & \best{61.3}  % I->T Flickr30K R@5
      & \best{72.4}  % I->T Flickr30K R@10
      & \best{5.851}  % ImageNet TIE (lower better)
      & \best{3.095}  % ImageNet LCA (lower better)
      & \best{0.641}  % ImageNet J (higher better)
      & \best{0.748}  % ImageNet P_H (higher better)
      & \best{0.747}  % ImageNet R_H (higher better)
      \\
    \cmidrule{2-16}
    & \multirow{2}{*}{ViT-B/16}
    & $\mathbb{E}^{n}$
      & \best{33.6}  % T->I COCO R@5
      & \best{44.8}  % T->I COCO R@10
      & \best{52.7}  % T->I Flickr30K R@5
      & \best{63.5}  % T->I Flickr30K R@10
      & \best{40.2}  % I->T COCO R@5
      & \best{52.4}  % I->T COCO R@10
      & 63.7  % I->T Flickr30K R@5
      & \best{75.2}  % I->T Flickr30K R@10
      & \best{5.514}  % ImageNet TIE (lower better)
      & \best{2.974}  % ImageNet LCA (lower better)
      & \best{0.664}  % ImageNet J (higher better)
      & \best{0.764}  % ImageNet P_H (higher better)
      & \best{0.764}  % ImageNet R_H (higher better)
      \\
    & & $\mathbb{L}^{n}_{c}$
      & 32.6  % T->I COCO R@5
      & 43.7  % T->I COCO R@10
      & 51.9  % T->I Flickr30K R@5
      & 63.1  % T->I Flickr30K R@10
      & 39.4  % I->T COCO R@5
      & 51.6  % I->T COCO R@10
      & \best{63.8}  % I->T Flickr30K R@5
      & 73.7  % I->T Flickr30K R@10
      & 5.719  % ImageNet TIE (lower better)
      & 3.103  % ImageNet LCA (lower better)
      & 0.652  % ImageNet J (higher better)
      & 0.753  % ImageNet P_H (higher better)
      & 0.757  % ImageNet R_H (higher better)
      \\
    \midrule
    \multirow{4}{*}{ARGENT}
    & \multirow{2}{*}{ViT-S/16}
    & $\mathbb{E}^{n}$
      & \best{30.4}  % T->I COCO R@5
      & 41.2  % T->I COCO R@10
      & \best{47.7}  % T->I Flickr30K R@5
      & \best{59.3}  % T->I Flickr30K R@10
      & \best{36.7}  % I->T COCO R@5
      & \best{48.2}  % I->T COCO R@10
      & \best{60.7}  % I->T Flickr30K R@5
      & \best{72.7}  % I->T Flickr30K R@10
      & 5.963  % ImageNet TIE (lower better)
      & \best{3.130}  % ImageNet LCA (lower better)
      & 0.634  % ImageNet J (higher better)
      & 0.743  % ImageNet P_H (higher better)
      & 0.740  % ImageNet R_H (higher better)
      \\
    & & $\mathbb{L}^{n}_{c}$
      & 30.1  % T->I COCO R@5
      & 41.2  % T->I COCO R@10
      & 46.7  % T->I Flickr30K R@5
      & 58.1  % T->I Flickr30K R@10
      & 35.8  % I->T COCO R@5
      & 47.7  % I->T COCO R@10
      & 57.6  % I->T Flickr30K R@5
      & 69.5  % I->T Flickr30K R@10
      & \best{5.931}  % ImageNet TIE (lower better)
      & 3.132  % ImageNet LCA (lower better)
      & \best{0.637}  % ImageNet J (higher better)
      & \best{0.746}  % ImageNet P_H (higher better)
      & \best{0.742}  % ImageNet R_H (higher better)
      \\
    \cmidrule{2-16}
    & \multirow{2}{*}{ViT-B/16}
    & $\mathbb{E}^{n}$
      & \best{33.1}  % T->I COCO R@5
      & \best{44.1}  % T->I COCO R@10
      & 52.3  % T->I Flickr30K R@5
      & \best{63.8}  % T->I Flickr30K R@10
      & \best{40.5}  % I->T COCO R@5
      & \best{52.9}  % I->T COCO R@10
      & \best{63.5}  % I->T Flickr30K R@5
      & 74.1  % I->T Flickr30K R@10
      & \best{5.579}  % ImageNet TIE (lower better)
      & \best{3.026}  % ImageNet LCA (lower better)
      & \best{0.661}  % ImageNet J (higher better)
      & \best{0.761}  % ImageNet P_H (higher better)
      & \best{0.762}  % ImageNet R_H (higher better)
      \\
    & & $\mathbb{L}^{n}_{c}$
      & 32.5  % T->I COCO R@5
      & 44.0  % T->I COCO R@10
      & \best{52.4}  % T->I Flickr30K R@5
      & 63.2  % T->I Flickr30K R@10
      & 39.5  % I->T COCO R@5
      & 51.7  % I->T COCO R@10
      & 62.9  % I->T Flickr30K R@5
      & \best{74.3}  % I->T Flickr30K R@10
      & 5.661  % ImageNet TIE (lower better)
      & 3.079  % ImageNet LCA (lower better)
      & 0.656  % ImageNet J (higher better)
      & 0.757  % ImageNet P_H (higher better)
      & 0.761  % ImageNet R_H (higher better)
      \\
    \midrule
    \multirow{4}{*}{PHyCLIP}
    & \multirow{2}{*}{ViT-S/16}
    & $\prod_{i=1}^{k}\mathbb{E}^{d}$
      & \best{30.6}  % T->I COCO R@5
      & \best{41.4}  % T->I COCO R@10
      & \best{48.8}  % T->I Flickr30K R@5
      & \best{59.7}  % T->I Flickr30K R@10
      & 37.2  % I->T COCO R@5
      & 48.6  % I->T COCO R@10
      & \best{59.2}  % I->T Flickr30K R@5
      & \best{72.6}  % I->T Flickr30K R@10
      & \best{5.920}  % ImageNet TIE (lower better)
      & \best{3.104}  % ImageNet LCA (lower better)
      & \best{0.636}  % ImageNet J (higher better)
      & \best{0.746}  % ImageNet P_H (higher better)
      & 0.742  % ImageNet R_H (higher better)
      \\
    & & $\prod_{i=1}^{k}\mathbb{L}^{d}_{c_i}$
      & 30.3  % T->I COCO R@5
      & 41.1  % T->I COCO R@10
      & 48.3  % T->I Flickr30K R@5
      & 59.2  % T->I Flickr30K R@10
      & \best{37.5}  % I->T COCO R@5
      & \best{49.2}  % I->T COCO R@10
      & 58.4  % I->T Flickr30K R@5
      & 70.6  % I->T Flickr30K R@10
      & 5.998  % ImageNet TIE (lower better)
      & 3.203  % ImageNet LCA (lower better)
      & 0.633  % ImageNet J (higher better)
      & 0.740  % ImageNet P_H (higher better)
      & 0.742  % ImageNet R_H (higher better)
      \\
    \cmidrule{2-16}
    & \multirow{2}{*}{ViT-B/16}
    & $\prod_{i=1}^{k}\mathbb{E}^{d}$
      & \best{33.4}  % T->I COCO R@5
      & \best{44.6}  % T->I COCO R@10
      & \best{52.9}  % T->I Flickr30K R@5
      & \best{64.3}  % T->I Flickr30K R@10
      & \best{40.3}  % I->T COCO R@5
      & \best{52.8}  % I->T COCO R@10
      & \best{64.4}  % I->T Flickr30K R@5
      & \best{75.0}  % I->T Flickr30K R@10
      & \best{5.588}  % ImageNet TIE (lower better)
      & \best{3.001}  % ImageNet LCA (lower better)
      & \best{0.660}  % ImageNet J (higher better)
      & \best{0.760}  % ImageNet P_H (higher better)
      & \best{0.760}  % ImageNet R_H (higher better)
      \\
    & & $\prod_{i=1}^{k}\mathbb{L}^{d}_{c_i}$
      & 32.8  % T->I COCO R@5
      & 44.0  % T->I COCO R@10
      & 51.8  % T->I Flickr30K R@5
      & 62.1  % T->I Flickr30K R@10
      & 39.6  % I->T COCO R@5
      & 52.1  % I->T COCO R@10
      & 62.8  % I->T Flickr30K R@5
      & 73.6  % I->T Flickr30K R@10
      & 5.745  % ImageNet TIE (lower better)
      & 3.122  % ImageNet LCA (lower better)
      & 0.652  % ImageNet J (higher better)
      & 0.752  % ImageNet P_H (higher better)
      & 0.755  % ImageNet R_H (higher better)
      \\
    \bottomrule
  \end{tabular}%
  }
\end{table}

\begin{figure}[H]
    \centering
    \includegraphics[width=\linewidth]{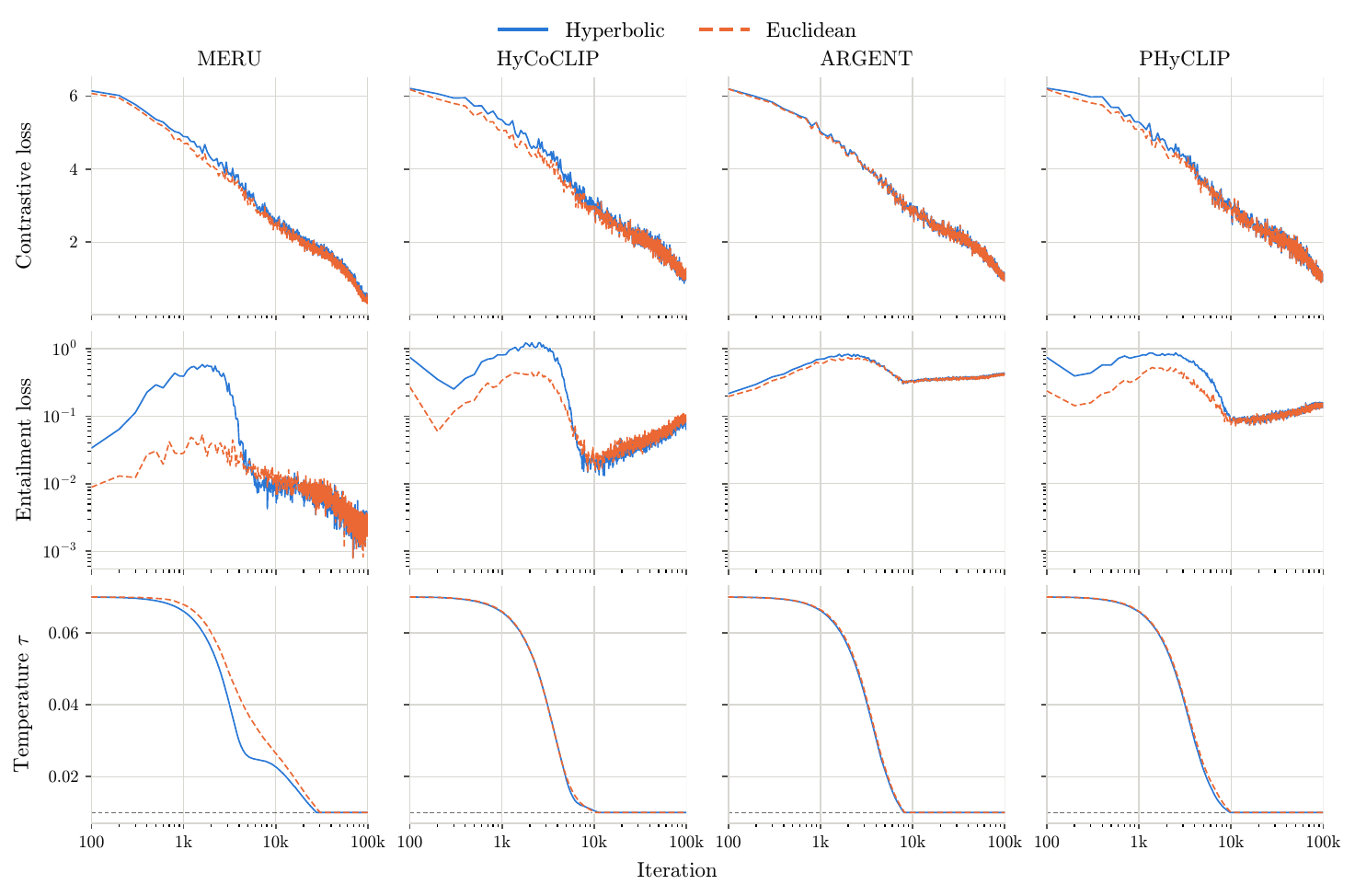}\vspace{-1cm}
    \caption{Training dynamics for the ViT-S/16 encoders, hyperbolic (solid) against Euclidean control (dashed). From top to bottom: the contrastive loss, the entailment loss on a logarithmic scale (the adaptive variant of \cref{eq:adaptive_entailment} for ARGENT), and the learned temperature $\tau$ with its lower clamp marked. The iteration axis is logarithmic throughout. The two runs separate over roughly the first $10{,}000$ steps, while the curvature falls to its floor, and track each other closely thereafter.}
    \label{fig:training_dynamics}
\end{figure}

\end{document}